\documentclass[manuscript,screen,acmsmall]{acmart}

\AtBeginDocument{%
  \providecommand\BibTeX{{%
    \normalfont B\kern-0.5em{\scshape i\kern-0.25em b}\kern-0.8em\TeX}}}

\setcopyright{acmcopyright}
\acmJournal{CSUR}
\acmYear{2021} \acmVolume{1} \acmNumber{1} \acmArticle{1} \acmMonth{1} \acmPrice{15.00}\acmDOI{10.1145/3494523}

\usepackage{amsfonts}
\usepackage{multirow}
\usepackage{tabularx}
\usepackage{makecell}
\usepackage{enumitem}
\begin{document}

\title{A Survey of Machine Learning for Computer Architecture and Systems}

\author{Nan Wu}
\email{nanwu@ucsb.edu}
\author{Yuan Xie}
\email{yuanxie@ucsb.edu}
\affiliation{%
  \institution{University of California, Santa Barbara}
  \city{Santa Barbara}
  \state{California}
  \postcode{93106}
}

\renewcommand{\shortauthors}{Wu and Xie.}

\begin{abstract}
It has been a long time that computer architecture and systems are optimized for efficient execution of machine learning (ML) models. 
Now, it is time to reconsider the relationship between ML and systems, and let ML transform the way that computer architecture and systems are designed.
This embraces a twofold meaning: improvement of designers' productivity, and completion of the virtuous cycle.
In this paper, we present a comprehensive review of the work that applies ML for computer architecture and system design.
First, we perform a high-level taxonomy by considering the typical role that ML techniques take in architecture/system design, i.e., either for fast predictive modeling or as the design methodology.
Then, we summarize the common problems in computer architecture/system design that can be solved by ML techniques, and the typical ML techniques employed to resolve each of them.
In addition to emphasis on computer architecture in a narrow sense, we adopt the concept that data centers can be recognized as warehouse-scale computers; sketchy discussions are provided in adjacent computer systems, such as code generation and compiler; we also give attention to how ML techniques can aid and transform design automation.
We further provide a future vision of opportunities and potential directions, and envision that applying ML for computer architecture and systems would thrive in the community.
\end{abstract}



\maketitle

\section{Introduction}
Machine learning (ML) has been doing wonders in many fields.
As people are seeking better artificial intelligence (AI), there is a trend towards larger, more expressive, and more complex models.
According to the data reported by OpenAI \cite{openai}, from 1959 to 2012, the amount of compute used in the largest AI training runs doubles every two years; since 2012, deep learning starts taking off, and the required amount of compute has been increasing exponentially with a 3.4-month doubling period. By comparison, Moore’s law \cite{moore1965cramming}, the principle that has powered the integrated-circuit revolution since 1960s, doubles the transistor density every 18 months.
While Moore's law is approaching its end \cite{waldrop2016chips}, more pressure is put on innovations of computer architecture and systems, so as to keep up with the compute demand of AI applications.

Conventionally, computer architecture/system designs are made by human experts based on intuitions and heuristics, which requires expertise in both ML and architecture/system.
Meanwhile, these heuristic-based designs can not guarantee scalability and optimality, especially in the case of increasingly complicated systems.
As such, it seems natural to move towards more automated and powerful methodologies for computer architecture and system design, and the relationship between ML and system design is being reconsidered.
Over the past decade, architecture and systems are optimized to accelerate the execution and improve the performance of ML models.
Recently, there have been signs of emergence of applying ML for computer architecture and systems, which embraces a twofold meaning: \textcircled{\small{1}} the reduction of burdens on human experts designing systems manually, so as to improve designers' productivity, and \textcircled{\small{2}} the close of the positive feedback loop, i.e., architecture/systems for ML and simultaneously ML for architecture/systems, formulating a virtuous cycle to encourage improvements on both sides.

\begin{figure}[tbp]
    \centering
    \vspace{-12pt}
	\includegraphics[width=0.95\linewidth]{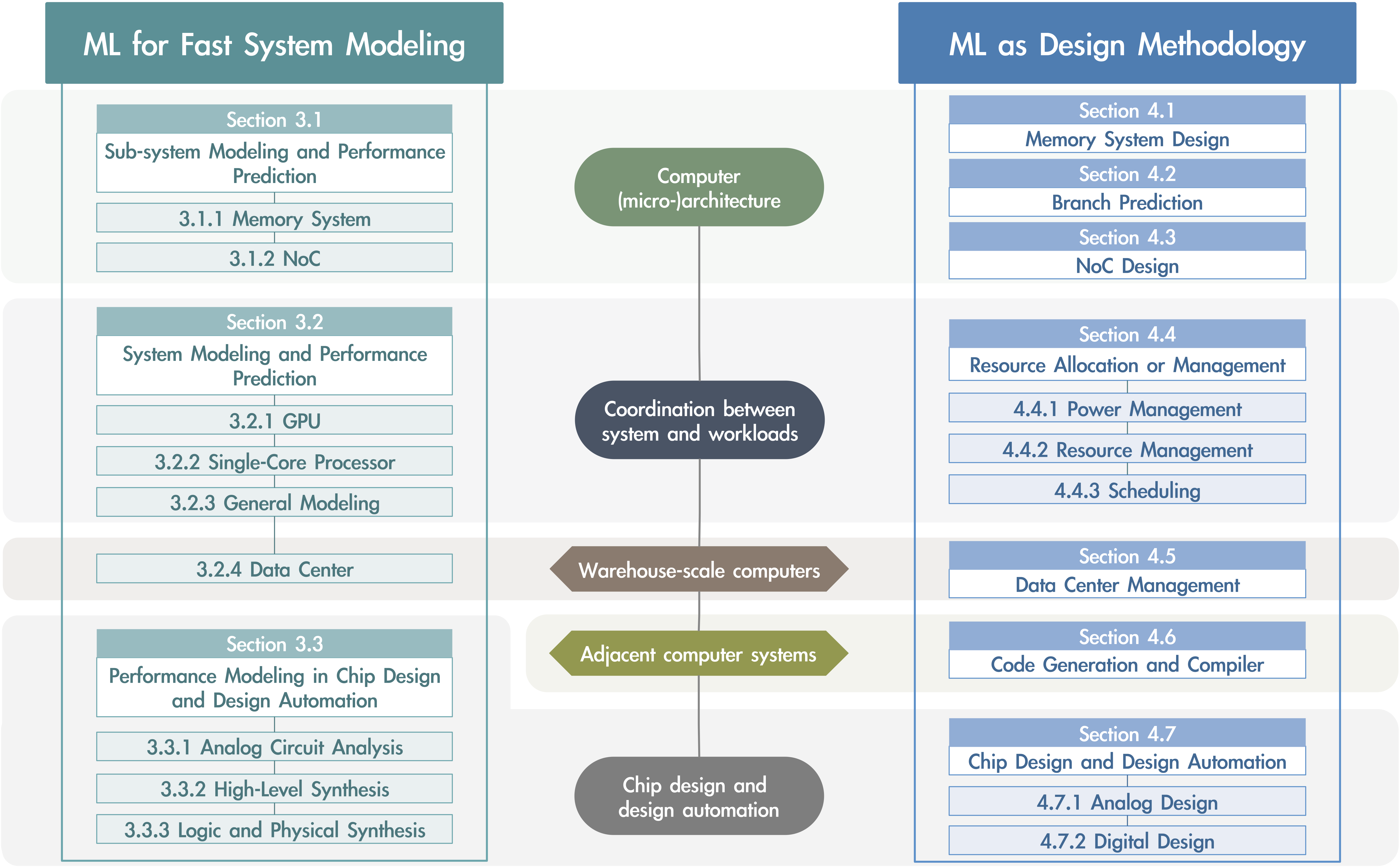}
	\vspace{-8pt}
    \caption{A comprehensive overview of applying ML for computer architecture and systems. Existing work roughly falls into two categories: ML for fast system modeling, and ML as design methodology.}
    \label{fig:frame}
    \vspace{-20pt}
\end{figure}

Existing work related to applying ML for computer architecture and system design falls into two categories.
\textcircled{\small{1}} ML techniques are employed for \textbf{fast and accurate system modeling}, which involves performance metrics or some criteria of interest (e.g. power consumption, latency, throughput, etc.).
During the process of designing systems, it is necessary to make fast and accurate predictions of system behaviors.
Traditionally, system modeling is achieved through the forms of cycle-accurate or functional virtual platforms, and instruction set simulators (e.g. gem5 \cite{binkert2011gem5}).
Even though these methods provide accurate estimations, they bring expensive computation costs associated with performance modeling, which limits the scalability to large-scale and complex systems; meanwhile, the long simulation time often dominates design iteration, making it impossible to fully explore the design space.
By contrast, ML-based modeling and performance prediction are capable to balance simulation cost and prediction accuracy.
\textcircled{\small{2}} ML techniques are employed as \textbf{a design methodology to directly enhance architecture/system design}.
ML techniques are skilled at extracting features that might be implicit to human experts, making decisions without explicit programming, and improving themselves automatically with accumulated experience.
Therefore, applying ML techniques as design tools can explore design space proactively and intelligently, and manage resource through better understanding of the complicated and non-linear interactions between workloads and systems, making it possible to deliver truly optimal solutions.

In this paper, we present a comprehensive overview of applying ML for computer architecture and systems. 
As depicted in Figure \ref{fig:frame},
we first perform a high-level taxonomy by considering the typical role that ML techniques take in architecture/system design, i.e., either for fast predictive modeling or as the design methodology;
then, we summarize the common problems in architecture/system design that can be solved by ML techniques, and the typical ML techniques employed to resolve each of them.
In addition to emphasis on computer architecture in a narrow sense, we adopt the concept that data centers can be recognized as warehouse-scale computers \cite{barroso2018datacenter}, and review studies associated with data center management; we provide sketchy discussions on adjacent computer systems, such as code generation and compiler; we also give attention to how ML techniques can aid and transform design automation that involves both analog and digital circuits.
At the end of the paper, we discuss challenges and future prospects of applying ML for architecture/system design, aiming to convey insights of design considerations.

\section{Different ML Techniques}
\label{sec:ml techniques}
There are three general frameworks in ML: supervised learning, unsupervised learning and reinforcement learning.
These frameworks mainly differentiate on what data are sampled and how these sample data are used to build learning models.
Table \ref{table:ml_techniques} summarizes the commonly used ML techniques for computer architecture and system designs.
Sometimes, multiple learning models may work well for one given problem, and the appropriate selection can be made based on available hardware resource and data, implementation overheads, performance targets, etc.

\begin{table}[tbp]
\vspace{-12pt}
\caption{Machine learning techniques.}
\vspace{-12pt}
\label{table:ml_techniques}
\centering
    \tiny
    \renewcommand{\arraystretch}{1}
    \setlength{\tabcolsep}{6pt}
\begin{tabular}{c|c|c|c}
\toprule
\textbf{Realm of ML} & \textbf{Category} & \makecell[c]{\textbf{Classical ML}} & \textbf{Deep Learning Counterpart \cite{goodfellow2016deep}} \\ \midrule
\multirow{10}{*}{\makecell*[c]{\textbf{Supervised} \\ \textbf{Learning}}} & \multirow{1}{*}{Classification} &
  Logistic regression \cite{hosmer2013applied} & 
  \multirow{8}{*}{\makecell{CNN, RNN,\\ GNN \cite{wu2020comprehensive}, etc.}} \\ \cline{2-3}
 & \multirow{6}{*}{Working for both} & Support vector machines/regression \cite{scholkopf2002learning} & \\ \cline{3-3}
 && K-nearest neighbors \cite{altman1992introduction} & \\ \cline{3-3}
 && Decision tree, e.g., CART \cite{loh2011classification}, MARS \cite{friedman1991multivariate} & \\ \cline{3-3}
 && ANN \cite{rumelhart1985learning} & \\ \cline{3-3}
 && Bayesian analysis \cite{gelman2013bayesian} & \\ \cline{3-3}
 && \makecell{Ensemble learning \cite{sagi2018ensemble}, e.g., gradient boosting, random forest} & \\ \cline{2-3}
 & \multirow{2}{*}{Regression} & \makecell{Linear regression with variants \cite{seber2012linear}, e.g., lasso (L1 regularization),\\ ridge (L2 regularization), elastic-net (hybrid L1/L2 regularization)} & \\ \cline{3-3}
 && Non-linear regression \cite{ritz2008nonlinear} & \\
 \hline
\multirow{2}{*}{\makecell[c]{\textbf{Unsupervised} \\ \textbf{Learning}}} 
& Clustering & K-means clustering \cite{jain2010data} &  \multirow{2}{*}{\makecell[c]{Autoencoder, \\ GAN, etc.}} \\ \cline{2-3}
 & Dimension reduction & Principal component analysis (PCA) \cite{wold1987principal} &\\ \hline
\multirow{3}{*}{\makecell*[c]{\textbf{Reinforcement} \\ \textbf{Learning}}} &
 Value-based & Q-learning \cite{sutton2018reinforcement} & DQN \cite{mnih2015human} \\ 
 \cline{2-4} 
 & \multirow{2}{*}{Policy-based} & Actor-critic \cite{sutton2018reinforcement}
 & A3C \cite{mnih2016asynchronous}, DDPG \cite{lillicrap2015continuous} \\ \cline{3-4}
 & & Policy gradient, e.g., REINFORCE \cite{sutton2018reinforcement}
 & PPO \cite{schulman2017proximal} \\
\bottomrule
\end{tabular}
\vspace{-15pt}
\end{table}

\vspace{-5pt}
\subsection{Supervised Learning}
\begin{figure}[tbp]
    \centering
    \vspace{-15pt}
	\includegraphics[width=0.97\linewidth]{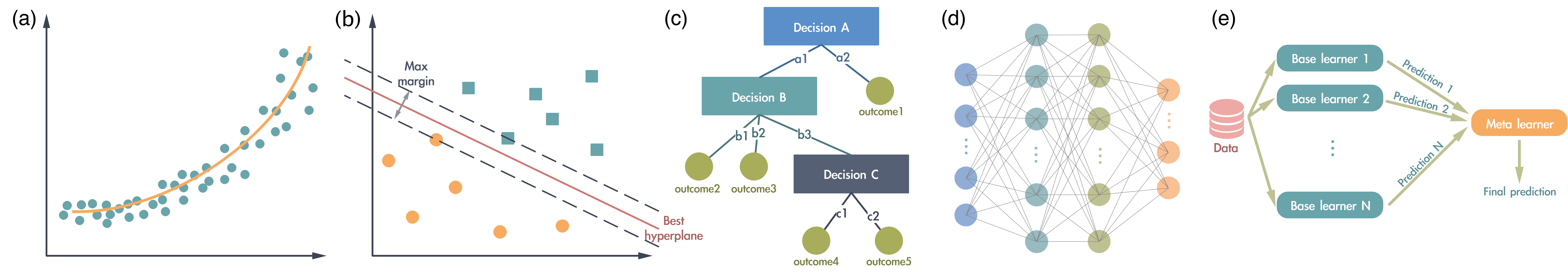}
	\vspace{-10pt}
    \caption{Examples of supervised learning: (a) regression, (b) SVM, (c) decision tree, (d) MLP, and (e) ensemble learning.}
    \vspace{-20pt}
    \label{fig:sl}
\end{figure}

Supervised learning is the process of learning a set of rules able to map an input to an output based on labeled datasets. These learned rules can be generalized to make predictions for unseen inputs.
We briefly introduce several prevalent techniques in supervised learning, as shown in Figure \ref{fig:sl}.

\begin{itemize}[leftmargin=*]
    \item Regression is a process for estimating the relationships between a dependent variable and one or more independent variables. The most common form is linear regression \cite{seber2012linear}, and some other forms include different types of non-linear regression \cite{ritz2008nonlinear}. Regression techniques are primarily used for two purposes, prediction/forecasting, and inference of causal relationships.
    \item Support vector machines (SVMs) \cite{scholkopf2002learning} try to find the best hyperplanes to separate data classes by maximizing margins. One variant is support vector regression (SVR), which is able to conduct regression tasks. Predictions or classifications of new inputs can be decided by their relative positions to these hyperplanes.
    \item Decision tree is one representative of logical learning methods, which uses tree structures to build regression or classification models.
    The final result is a tree with decision nodes and leaf nodes. Each decision node represents a feature and branches of this node represent possible values of the corresponding feature. Starting from the root node, input instances are classified by sequentially passing through nodes and branches, until they reach leaf nodes that represent either classification results or numerical values.
    \item Artificial neural networks (ANNs) \cite{rumelhart1985learning} are capable to approximate a broad family of functions: a single-layer perceptron is usually used for linear regression; complex DNNs \cite{goodfellow2016deep} consisting of multiple layers are able to approximate non-linear functions, such as the multi-layer perceptron (MLP); variants of DNNs that achieve excellent performance in specific fields benefit from the exploitation of certain computation operations, e.g., convolutional neural networks (CNNs) with convolution operations leveraging spatial features, and recurrent neural networks (RNNs) with recurrent connections enabling learning from sequences and histories.
    \item Ensemble learning \cite{sagi2018ensemble} employs multiple models that are strategically designed to solve a particular problem, and the primary goal is to achieve better predictive performance than those could be obtained from any of the constituent models alone. Several common types of ensembles include random forest and gradient boosting.
\end{itemize}

Different learning models have different preference of input features: 
SVMs and ANNs generally perform much better with multi-dimension and continuous features, while logic-based systems tend to perform better when dealing with discrete/categorical features.
In system design, supervised learning is commonly used for performance modeling, configuration predictions, or predicting higher-level features/behaviors from lower-level features.
One thing worth noting is that supervised learning techniques need well labeled training data prior to the training phase, which usually require tremendous human expertise and engineering.

\vspace{-5pt}
\subsection{Unsupervised Learning}
\begin{figure}[t]
\vspace{-15pt}
\flushleft
\begin{minipage}{.63\textwidth}
  \centering
  \includegraphics[width=0.82\linewidth]{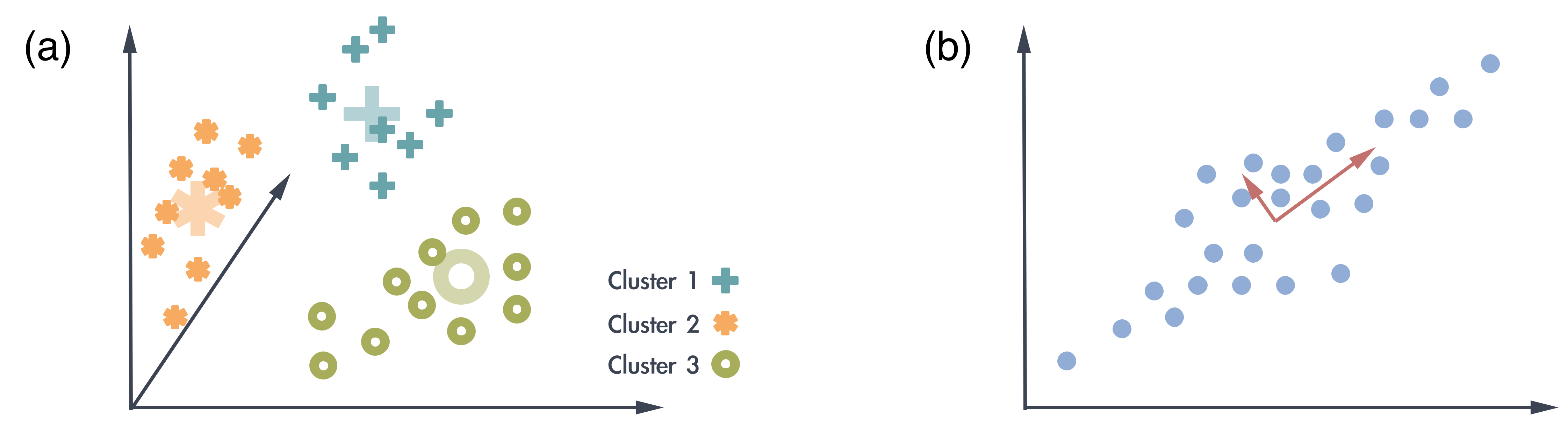}
  \vspace{-10pt}
  \caption{Examples of unsupervised learning: (a) clustering, and (b) PCA.}
  \label{fig:usl}
\end{minipage}%
\hspace{.03\textwidth}
\begin{minipage}{.33\textwidth}
  \centering
  \includegraphics[width=\linewidth]{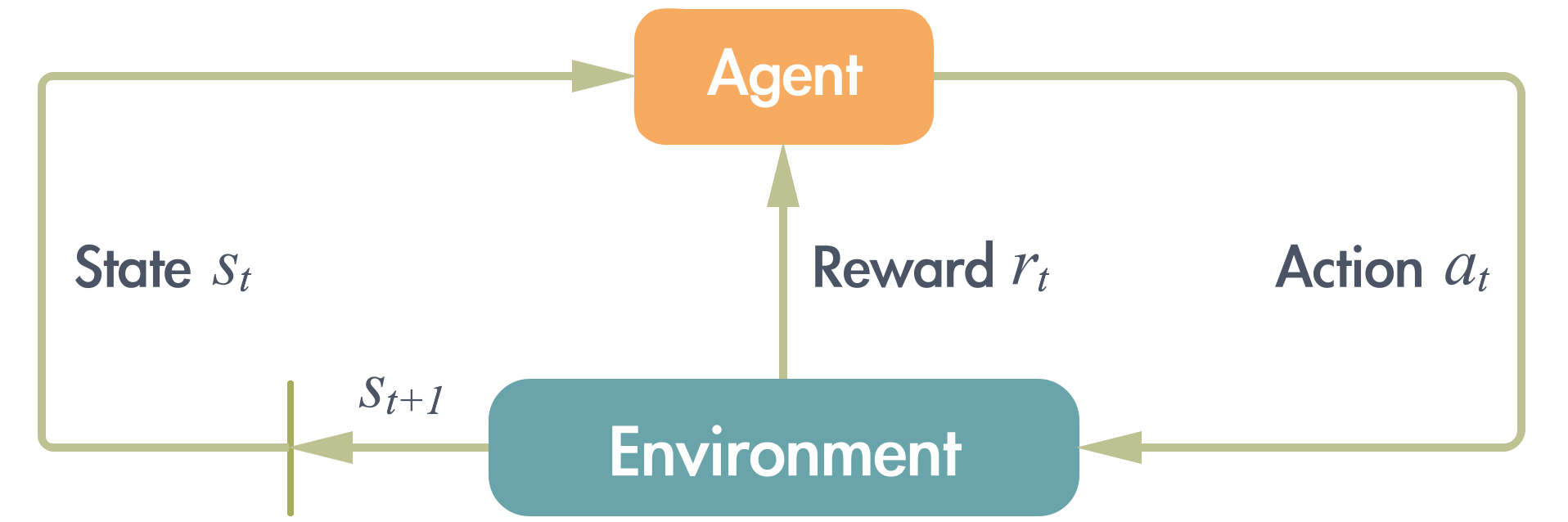}
  \caption{A typical framing of RL.}
  \label{fig:rl}
\end{minipage}%
\vspace{-10pt}
\end{figure}

Unsupervised learning is the process of finding previously unknown patterns based on unlabeled datasets.
Two prevailing methods are clustering analysis \cite{jain2010data} and principal component analysis (PCA) \cite{wold1987principal}, as depicted in Figure \ref{fig:usl}.

\begin{itemize}[leftmargin=*]
    \item Clustering is a process of grouping data objects into disjoint clusters based on a measure of similarity, such that data objects in the same cluster are similar while data objects in different clusters share low similarities. The goal of clustering is to classify raw data reasonably and to find possibly existing hidden structures or patterns in datasets. One of the most popular and simple clustering algorithms is k-means clustering.
    \item PCA is essentially a coordinate transformation leveraging information from data statistics. It aims to reduce the dimensionality of the high-dimensional variable space by representing it with a few orthogonal (linearly uncorrelated) variables that capture most of its variability.
\end{itemize}

Since there is no label in unsupervised learning, it is difficult to simultaneously measure the performance of learning models and decide when to stop the learning process.
One potential workaround is \textit{semi-supervised learning} \cite{zhu2005semi}, which uses a small amount of labeled data together with a large amount of unlabeled data. This approach stands between unsupervised and supervised learning, requiring less human effort and producing higher accuracy.
The unlabeled data are used to either finetune or re-prioritize hypotheses obtained from labeled data alone.

\subsection{Reinforcement Learning}

In standard reinforcement learning (RL) \cite{sutton2018reinforcement}, an agent interacts with an environment $\mathcal{E}$ over a number of discrete time steps, as shown in Figure \ref{fig:rl}.
At each time step $t$, the agent receives a state $s_t$ from the \textit{state space} $\mathcal{S}$, and selects an action $a_t$ from the \textit{action space} $\mathcal{A}$ according to its policy $\pi$, where $\pi$ is a mapping from states $s_t$ to actions $a_t$. 
In return, the agent receives the next state $s_{t+1}$ and a scalar reward $r_t: \mathcal{S}\times\mathcal{A}\to\mathbb{R}$. 
This process continues until the agent reaches a terminal state after which the process restarts.
The return $R_t={\sum\limits_{k=0}^\infty \gamma^kr_{t+k}}$ is the totally accumulated rewards at the time step $t$ with a \textit{discount factor} $\gamma\in(0,1]$. 
The goal of the agent is to maximize the expected return for each state $s$.

The state-action value $Q_{\pi}(s,a)=\mathbb{E_\pi}\lbrack R_t|s_t=s, a_t=a\rbrack$ is the expected return of selecting action $a$ at state $s$ with policy $\pi$.
Similarly, the state value $V_{\pi}(s)=\mathbb{E_\pi}\lbrack R_t|s_t=s\rbrack$ is the expected return starting from state $s$ by following policy $\pi$. There are two general types of methods in RL: value-based, and policy-based.

\begin{itemize}[leftmargin=*]
    \item In value-based RL, the state-action value function $Q_{\pi}(s,a)$ is approximated by either tabular approaches or function approximations. At each state $s_t$, the agent always selects the optimal action $a^*_t$ that could bring the maximal state-action value $Q_{\pi}(s_t,a^*_t)$. One well-known example of value-based methods is Q-learning.
    \item In policy-based RL, it directly parameterizes the policy $\pi(a|s;\theta)$ and updates the parameters $\theta$ by performing gradient ascent on $\mathbb{E}[R_t]$. One example is the REINFORCE algorithm. 
\end{itemize}

RL is modeled based on Markov decision process, and thus it is suitable to handle control problems or sequential decision-making processes.
With these characteristics, RL is able to explore design space proactively and intelligently, and learn how to achieve resource management or task scheduling in system designs through interactions with environments. The optimal behaviors can be found by embedding optimization goals into reward functions.

\section{ML for Fast System Modeling}

\label{sec:system modeling}

This section reviews studies that employ ML techniques for fast and accurate system modeling, which involves predictions of performance metrics or some other criteria of interest.
Although cycle-accurate simulators, which are commonly used for system performance prediction, can provide accurate estimations, they usually run multiple orders of magnitude slower than native executions.
By contrast, ML-based techniques can balance simulation costs and prediction accuracy, showing great potentials in exploring huge configuration spaces and learning non-linear impacts of configurations.
Most of existing work applies supervised learning for either pure system modeling or efficient design space exploration (DSE) enabled by fast predictions.
Table \ref{table:model_sys} and Table \ref{table:model_eda} summarize the studies for predictive modeling in computer architecture/system and design automation respectively, in terms of task domains, prediction targets, adopted ML techniques, and corresponding inputs.

\begin{table}[tbp]
\vspace{-10pt}
\caption{Summary of applying ML techniques for fast modeling in computer architecture and systems.}
\label{table:model_sys}
\centering
    \tiny
    \renewcommand{\arraystretch}{0.9}
    \setlength{\tabcolsep}{1pt}
\vspace{-10pt}
\begin{tabular}{c|c|c|c}
\toprule
\textbf{Domain}  & \textbf{Prediction of}  & \textbf{Technique} & \textbf{Input} \\ \midrule
\multirow{4}{*}{\makecell[c]{\textbf{Memory} \\ \textbf{System} \\ $( \S$ \textbf{\ref{sec:model_memory}}$)$}} 
& Throughput/cache miss & ANN \cite{dong2013circuit}  & Cache configurations  \\ \cline{2-4} 
& Throughput/lifetime/energy & \makecell[c]{Gradient boosting/quadratic regression with lasso \cite{deng2017memory}}  & NVM configurations  \\ \cline{2-4} 
& Throughput   & CNN \cite{lin2019design}   & Memory controller placements   \\ \cline{2-4} 
& Disk block correlation  & DNN \cite{dai2016block2vec} & Data blocks in the context window \\ \hline
\multirow{4}{*}{\makecell[c]{\textbf{NoC} \\ $( \S$ \textbf{\ref{sec:model_noc}}$)$}}           
& Latency  & SVR \cite{qian2013svr} & Queue information \\ \cline{2-4} 
& Hotspots & ANN \cite{soteriou2015holistic} & Buffer utilization \\ \cline{2-4} 
& Buffer/link utilization  & \makecell[c]{Regression with ridge regularization \cite{clark2018lead,van2018extending}, decision tree \cite{ditomaso2017machine}} & NoC configurations  \\ \cline{2-4} 
& Probability of errors  & Decision tree \cite{ditomaso2016dynamic}                     & \makecell[c]{Link utilization and transistor wearing-out} \\ \hline
\multirow{6}{*}{\makecell[c]{\textbf{GPU} \\ $( \S$ \textbf{\ref{sec:model_gpu}}$)$}} & \makecell[c]{Speedup or execution time \\ by cross-platform inputs} & \makecell[c]{Nearest neighbor/SVM \cite{baldini2014predicting}, \\ ensemble of regression-based learners \cite{ardalani2015cross}, \\ random forest \cite{ardalani2019static}} & \makecell[c]{Static analysis and/or \\ dynamic profiling of \\ source CPU code} \\ \cline{2-4} 
 & Execution time & \makecell[c]{Stepwise regression \cite{jia2012stargazer}, \\ ensemble of linear regression and random forest \cite{o2017halwpe}} & \multirow{3}{*}{\makecell[c]{GPU configurations \\ and performance counters}} \\ \cline{2-3}
 & Throughput/power & Ensemble of NNs \cite{jooya2016multiobjective} &  \\ \cline{2-3}
 & Scaling behavior of GPGPU & ANN and K-means clustering \cite{wu2015gpgpu} &  \\ \cline{2-4} 
 & Kernel affinity and execution time & Logistic regression and linear regression \cite{pattnaik2016scheduling} & Kernel characteristics \\ \cline{2-4}
 & Traffic patterns in GPGPU & CNN \cite{li2019characterizing} & Grayscale heat maps \\ \hline
\multirow{2}{*}{\makecell[c]{\textbf{Single-Core} \\ \textbf{CPU} $( \S$ \textbf{\ref{sec:model_single}}$)$}} 
& Throughput & \makecell[c]{Linear regression \cite{joseph2006construction}, non-linear regression   \cite{joseph2006predictive,lee2006accurate,eyerman2011mechanistic}} & \multirow{2}{*}{\makecell[c]{Micro-architectural parameters \\ and performance counters}} \\ \cline{2-3}
 & Program execution time & Linear   regression\cite{zheng2015learning,zheng2016accurate} &  \\ \hline
\multirow{8}{*}{\makecell[c]{\textbf{General}\\ \textbf{Modeling} \\ $( \S$ \textbf{\ref{sec:model_general}}$)$}} & \multirow{6}{*}{Throughput/latency/power} & ANN \cite{ipek2006efficiently,khan2007using, lee2007methods, ozisikyilmaz2008machine,bitirgen2008coordinated,nemirovsky2017machine} & \multirow{7}{*}{\makecell[c]{Micro-architectural parameters \\ and performance counters}} \\ \cline{3-3}
 &  & \makecell[c]{Non-linear regression  \cite{lee2007illustrative,lee2008cpr,wu2012inferred,sangaiah2015uncore}} &  \\ \cline{3-3}
 &  & Linear regression \cite{curtis2008prediction,bailey2014adaptive,lo2015prediction} &  \\ \cline{3-3}
 &  & Hierarchical Bayesian model \cite{mishra2015probabilistic,mishra2018caloree} &  \\ \cline{3-3}
 &  & LSTM   \cite{mendis2019ithemal,renda2020difftune} &  \\ \cline{3-3}
 &  & Generative model \cite{ding2019generative} &  \\ \cline{2-3}
 & \makecell[c]{Slowdown caused by \\ application interference} & \makecell[c]{Linear regression with elastic-net \\ regularization \cite{mishra2017esp}} &  \\ \cline{2-4} 
 & \makecell[c]{Speedup of multi-thread applications} & Gausian process regression \cite{agarwal2019performance} & Profiling of single-thread execution \\ \hline
\multirow{9}{*}{\makecell[c]{\textbf{Data} \\ \textbf{Center} \\ $( \S$ \textbf{\ref{sec:model_dc}}$)$}} & Job completion time & SVR \cite{yigitbasi2013towards} & \makecell[c]{Application characteristics \\ and cluster configurations} \\ \cline{2-4} 
 & Resource demand & \makecell[c]{Statistical learning \cite{gong2010press}, linear regression/MLP \cite{islam2012empirical}} & \multirow{3}{*}{Workload characterization} \\ \cline{2-3}
 & Incoming workload & ARMA \cite{roy2011efficient},   ARIMA \cite{calheiros2014workload} &  \\ \cline{2-3}
 & Workload pattern & Hidden Markov model \cite{khan2012workload} &  \\ \cline{2-4} 
 & Power   usage effectiveness & MLP \cite{gao2014machine} & Data center configurations \\ \cline{2-4} 
 & Disk Failure & \makecell[c]{Bayesian methods \cite{hamerly2001bayesian}, clustering \cite{murray2005machine}, \\ SVM/MLP \cite{zhu2013proactive}, random forest \cite{xiao2018disk}} & \multirow{3}{*}{\makecell[c]{SMART (Self-Monitoring, Analysis \\ and Reporting Technology) attributes \\ of data centers}} \\ \cline{2-3}
 & Health assessment of drives & \makecell[c]{CART \cite{li2014hard}, gradient boosted regressions tree \cite{li2017hard}, RNN \cite{xu2016health}} &  \\ \cline{2-3}
 & \multirow{2}{*}{Partial drive failure} & \makecell[c]{CART/random forest/SVM/ANN/logistic regression \cite{mahdisoltani2017proactive}} &  \\ \cline{3-4} 
 &  & Gradient boosted regression trees \cite{xu2018improving} & \makecell[c]{SMART attributes and system-level signals} \\ \bottomrule
\end{tabular}
\vspace{-15pt}
\end{table}

\subsection{Sub-system Modeling and Performance Prediction}
\label{sec:model_sub_sys}

\subsubsection{Memory System}
\label{sec:model_memory}
In memory systems, ML-based performance models are exploited to help explore trade-offs among different objectives.
To explore non-volatile memory (NVM) based cache hierarchies, Dong \textit{et al.} \cite{dong2013circuit} develop an ANN model to predict higher-level features (e.g. miss of cache read/write, and instruction-per-cycle (IPC)) from lower-level features (e.g. cache associativity, capacity and latency). 
To adaptively select architectural techniques in NVMs for different applications, Memory Cocktail Therapy \cite{deng2017memory} estimates lifetime, IPC, and energy consumption through lightweight online predictors by gradient boosting and quadratic regression with lasso. 
To optimize memory controller placements in throughput processors, Lin \textit{et al.} \cite{lin2019design} build a CNN model that takes memory controller placements as inputs to predict throughput, which accelerates the optimization process by two orders of magnitude.

Some studies concentrate on learning efficient representations of memory access patterns.
Block2Vec \cite{dai2016block2vec} tries to mine data block correlations by training a DNN to learn the best vector representation of each block and capturing block similarities via vector distances, which enables further optimization for caching and prefetching.
Shi \textit{et al.} \cite{shi2019learning} use a graph neural network (GNN) to learn fused representations of static code and its dynamic execution. This unified representation is capable to model both data flows (e.g., prefetching) and control flows (e.g., branch prediction).

\subsubsection{Network-on-Chip (NoC)}
\label{sec:model_noc}
In NoCs, several performance metrics of interest are latency, energy consumption, and reliability.
\textcircled{\small{1}}
Regarding latency predictions, Qian \textit{et al.} \cite{qian2013svr} use an SVR model to predict the traffic flow latency and the average channel waiting time in mesh-based NoCs, which relaxes some assumptions in the classical queuing theory. 
Rather than explicitly predicting latency, a lightweight hardware-based ANN \cite{soteriou2015holistic} predicts existence of traffic hotspots, which are intensive network congestions significantly degrading the effective throughput and implicitly indicate the average communication latency in NoCs.
The input features are buffer utilization rates from neighboring NoC routers, and the trained predictor is combined with a proactive hotspot-preventive routing algorithm to avert hotspot formation, attaining significant improvements for synthetic workloads while modest melioration for real-world benchmarks.
\textcircled{\small{2}}
Regarding estimating energy consumption, the learned predictors are often leveraged for saving dynamic and/or static energy in NoCs.
DiTomaso \textit{et al.} \cite{ditomaso2017machine} use per-router decision trees to predict link utilization and traffic direction, which are combined with sleepy link storage units to power-gate links/routers and to change link directions. 
Clark \textit{et al.} \cite{clark2018lead} use ridge regression models to predict buffer utilization, changes in buffer utilization, or a combined metric of energy and throughput,
based on which a router can select proper voltage/frequency.
In photonic NoCs, the ridge regression model is also applicable to predict the number of packets to be injected into each router in the following time window \cite{van2018extending}, based on which the number of wavelengths are scaled properly to reduce static energy consumed by photonic links.
\textcircled{\small{3}}
Regarding the reliability of NoCs, a per-link decision tree trained offline can predict the probability of timing faults on links during runtime \cite{ditomaso2016dynamic}, based on which a proactive fault-tolerant technique is developed to mitigate errors by using the strengthened cyclic redundancy check with error-correction code and relaxed transmission.

\subsection{System Modeling and Performance Prediction}
\label{sec:model_sys}
Accurate and fast performance estimation is a necessity for system optimization and design space exploration.
With the increasing complexity of systems and variety of workloads, ML-based techniques can provide highly accurate performance estimations with reasonable simulation costs, surpassing the capability of commonly-used cycle-accurate simulators that require highly computational costs and long simulation time.

\subsubsection{Graphics Processing Unit (GPU)}
\label{sec:model_gpu}
There are two types of predictions for GPU modeling: cross-platform predictions and GPU-specific predictions.
Cross-platform predictions are used to decide in advance whether to offload an application from a CPU to a GPU, since not every application benefits from GPU execution and the porting process requires considerably additional efforts;
GPU-specific predictions are used to estimate metrics of interest and to assist GPU design space exploration, helpful to handle design space irregularities and complicated interactions among configurations.

Cross-platform predictions can be formulated as a binary classification problem that identifies whether the potential GPU speedup of an application would be greater than a given threshold.
This task can be solved by the nearest neighbor and SVMs using dynamic instruction profiles  \cite{baldini2014predicting}, or a random forest that composes of one thousand decision trees using static analysis of source CPU code (i.e., memory coalescing, branch divergence, kernel size available parallelism and instruction intensities) \cite{ardalani2019static}.
With both dynamic and static program properties from single-thread CPU code, an ensemble of one hundred regression-based learners can predict the GPU execution time \cite{ardalani2015cross}.

In terms of GPU-specific predictions that take GPU configurations and performance counters as input features, the execution time can be predicted by stepwise linear regression, which recognizes the most important input features among many GPU parameters and thus achieves high accuracy even with sparse samples \cite{jia2012stargazer}; the power/throughput can be modeled by an ensemble of NN predictors \cite{jooya2016multiobjective}.
Provided with profiling results from earlier-generation GPUs, an ensemble of linear and non-linear regression models is capable to predict cross-generation GPU execution time for later/future-generation GPUs, which achieves more than 10,000 times speedup compared to cycle-accurate GPU simulators \cite{o2017halwpe}.
Focusing on processing-in-memory (PIM) assisted GPU architectures, Pattnaik \textit{et al.} \cite{pattnaik2016scheduling} classify GPU cores into two types: powerful GPU cores but far away from memory, and auxiliary/simple GPU cores but close to memory. They develop a logistic regression model that takes kernel characteristics as input features to predict architecture affinity of kernels, aiming to accurately identify which kernels would benefit from PIM and offload them accordingly to auxiliary GPU cores. They also build a linear regression model to predict the execution time of each kernel, so that a concurrent kernel management mechanism can be developed based on these two models and kernel dependency information. 
Focusing on general-purpose GPUs (GPGPUs), Wu \textit{et al.} \cite{wu2015gpgpu} model kernel scaling behaviors with respect to the number of compute units, engine frequency, and memory frequency. During training, kernels with similar performance scaling behaviors are grouped by K-means clustering, and when encountering a new kernel, it is mapped to the cluster that best describes its scaling performance by an ANN-based classifier.
Li \textit{et al.} \cite{li2019characterizing} reassess prevailing assumptions of GPGPU traffic patterns, and combine a CNN with a t-distributed stochastic neighbor embedding to classify different traffic patterns.

\subsubsection{Single-Core Processor}
\label{sec:model_single}
In predictive performance modeling of single-core processors, early-stage work mostly targets superscalar processors. 
To predict the application-specific cycle-per-instruction (CPI) of superscalar processors, Joseph \textit{et al.} \cite{joseph2006construction} introduce an iterative procedure to build linear regression models using 26 key micro-architectural parameters. Later they construct predictive models by non-linear regression techniques (i.e., radial basis function networks generated from regression trees) with 9 key micro-architectural parameters \cite{joseph2006predictive}.
In parallel with Joseph's work, Lee and Brooks \cite{lee2006accurate} use regression modeling with cubic splines to predict application-specific performance (billions of instructions per second) and power.

Later work focuses on performance modeling for existing hardware (e.g., Intel, AMD, and ARM processors) by using micro-architectural parameters and performance counters.
Eyerman \textit{et al.} \cite{eyerman2011mechanistic} construct a mechanistic-empirical model for CPI predictions of three Intel processors. The initially parameterized performance model is inspired by mechanistic modeling, where the unknown parameters inside the model are derived through regression, benefiting from both mechanistic modeling (i.e., interpretability) and empirical modeling (i.e., ease of implementation).
Zheng \textit{et al.} \cite{zheng2015learning,zheng2016accurate} explore two approaches to cross-platform predictions of program execution time, where profiling results on Intel Core i7 and AMD Phenom processors are used to estimate the execution time on a target ARM processor. The first approach \cite{zheng2015learning} relaxes the assumption of global linearity to local linearity in the feature space and applies constrained locally sparse linear regression; the other approach \cite{zheng2016accurate} applies lasso linear regression with phase-level performance features.

\subsubsection{General Modeling and Performance Prediction}
\label{sec:model_general}
Regression techniques are the mainstream to predict performance metrics from micro-architectural parameters or other features, which attributes to their capability to make high-accuracy estimations with reasonable training costs.

For conventional regression-based models, ANNs and non-linear regression with different designs are the common practice to predict throughput/latency \cite{ipek2006efficiently,lee2007illustrative,khan2007using,lee2008cpr,wu2012inferred} and power/energy \cite{khan2007using,lee2007illustrative}.
Consequently, there are comparisons among different techniques.
Lee \textit{et al.} \cite{lee2007methods} compare piecewise polynomial regression with ANNs, with emphasis that piecewise polynomial regression offers better explainability while ANNs show better generalization ability.
Ozisikyilmaz \textit{et al.} \cite{ozisikyilmaz2008machine} contrast several linear regression models and different ANNs, indicating that the pruned ANNs achieve best accuracy while requiring longer training time.
Agarwal \textit{et al.} \cite{agarwal2019performance} estimate the parallel execution speedup of multi-threaded applications on a target hardware platform, and mention that Gaussian process regression performs the best among several explored methods in this case.

More recent work tends to take advantage of data-driven approaches.
Ithemal \cite{mendis2019ithemal} leverages a hierarchical multi-scale RNN with long short term memory (LSTM) to predict throughput of basic blocks (i.e., sequences of instructions with no branches or jumps), and evaluations demonstrate that Ithemal is more accurate and as fast as analytical throughput estimators.
By employing a variant of Ithemal as a differentiable surrogate to approximate CPU simulators, DiffTune \cite{renda2020difftune} is able to apply gradient-based optimization techniques to learn the parameters of x86 basic block CPU simulators such that simulators' error is minimized. The learned parameters finally are plugged back into the original simulator.
Ding \textit{et al.} \cite{ding2019generative} provide some insights in learning-based modeling methods: the improvement of prediction accuracy may receive diminishing returns; the consideration of domain knowledge will be helpful for system optimizations, even if the overall accuracy may not be improved.
Thus, they propose a generative model to handle data scarcity by generating more training data, and apply a multi-phase sampling to improve prediction accuracy of optimal configuration points.

ML-based predictive performance modeling enables efficient resource management and rapid design space exploration to improve throughput.
Equipped with ANNs for IPC predictions, strategies for resource allocation \cite{bitirgen2008coordinated} and task scheduling \cite{nemirovsky2017machine} can always select decisions that would bring the best predicted IPC.
ESP \cite{mishra2017esp} constructs a regression model with elastic-net regularization to predict application interference (i.e., slowdown), which is integrated with schedulers to increase throughput.
MetaTune \cite{ryu2021metatune} is a meta-learning based cost model for convolution operations, and when combined with search algorithms, it enables efficient auto-tuning of parameters during compilation.
In consideration of rapid design space exploration of the uncore (i.e., memory hierarchies and NoCs), Sangaiah \textit{et al.} \cite{sangaiah2015uncore} uses a regression-based model with restricted cubic splines to estimate CPI, reducing the exploration time by up to four orders of magnitude.

ML-based predictive performance modeling benefits adaptations between performance and power budgets.
Leveraging off-line multivariate linear regression to predict IPC and/or power of different architecture configurations, Curtis-Maury \textit{et al.} \cite{curtis2008prediction} maximize performance of OpenMP applications by dynamic concurrency throttling and dynamic voltage and frequency scaling (DVFS); Bailey \textit{et al.} \cite{bailey2014adaptive} apply hardware frequency-limiting techniques to select optimal hardware configurations under given power constraints.
To effectively apply DVFS towards various optimization goals, the designed strategy can adopt predictions for power consumption by a constrained-posynomial model \cite{juan2013learning} or job execution time by a linear regression model \cite{lo2015prediction}.
To conduct smart power management in a more general manner, LEO \cite{mishra2015probabilistic} employs hierarchical Bayesian models to predict performance and power, and when integrated for runtime energy optimization, it is capable to figure out the performance-power Pareto frontier and select the configuration satisfying performance constraints with minimized energy.
CALOREE~\cite{mishra2018caloree} further breaks up the power management task into two abstractions: a learner for performance modeling and an adaptive controller leveraging predictions from the learner. These abstractions enable both the learner to use multiple ML techniques and the controller to maintain control-theoretic formal guarantees. Since no user-specified parameter except the goal is required, CALOREE is applicable even for non-experts.

\subsubsection{Data Center Performance Modeling and Prediction}
\label{sec:model_dc}
Data centers have been in widespread use for both traditional enterprise applications and cloud services.
Many studies employ ML techniques to predict workload/resource-related metrics, so as to enable elastic resource provision.
Common examples include but not limited to using SVR to predict job completion time \cite{yigitbasi2013towards}, leveraging the autoregressive moving average (ARMA) model \cite{roy2011efficient} or the autoregressive integrated moving average (ARIMA) model \cite{calheiros2014workload} to forecast incoming workloads, exploiting the hidden Markov modeling to characterize variations in workload patterns \cite{khan2012workload}, and estimating dynamic resource demand of workloads by light-weight statistical learning algorithms \cite{gong2010press} or MLP \cite{islam2012empirical}.
Jim Gao \cite{gao2014machine} builds an MLP model to predict power usage effectiveness of data centers, which is extensively tested and validated at Google data centers.
Cortez \textit{et al.} \cite{cortez2017resource} predict virtual machine (VM) behaviors (including VM lifetimes, maximum deployment sizes, and workload classes) for a broader set of purposes (e.g., health/resource management and power capping), where the evaluated ML models are random forests and extreme gradient boosting trees.

In addition to workload/resource-related metrics, the availability in data centers or cloud services is also a topic of concern, where one of the key tasks is to predict disk failure in advance.
Leveraging SMART (Self-Monitoring, Analysis and Reporting Technology) attributes, the disk failure prediction model can be built via various ML techniques, such as different Bayesian methods \cite{hamerly2001bayesian}, unsupervised clustering \cite{murray2005machine}, SVM and MLP \cite{zhu2013proactive}.
The adoption of classification and regression trees (CART) \cite{li2014hard}, RNNs \cite{xu2016health}, or gradient boosted regression trees \cite{li2017hard} makes it possible to assess health status of drives.
While all the aforementioned methods rely on offline training, online random forests \cite{xiao2018disk} can evolve with forthcoming data on-the-fly by generating new trees and forget old information by discarding outdated trees, consequently avoiding the model aging problem in disk failure predictions.
To predict partial drive failures (i.e., disk error or sector error), Mahdisoltani \textit{et al.} \cite{mahdisoltani2017proactive} explore five ML techniques (CART, random forests, SVM, NN and logistic regression), among which random forests consistently outperform others.
Xu \textit{et al.} \cite{xu2018improving} incorporate SMART attributes and system-level signals to train a gradient boosted regression tree, which is an online prediction model that ranks disks according to the degree of error-proneness in the near future.

\subsection{Performance Modeling in Chip Design and Design Automation}
\label{sec:model_eda}

\subsubsection{Analog Circuit Analysis}
\label{sec:model_analog}
Analog circuit design is usually a manual process that requires many trial-and-error iterations between pre-layout and post-layout phases.
In recent years, the discrepancy between schematic (i.e., pre-layout) performance estimations and post-layout simulation results is further enlarged.
On the one hand, the analytical performance estimations from schematics are no longer accurate with device scaling; on the other hand, even though post-layout simulations can provide high-accuracy estimations, they are extremely time-consuming and have become the major bottleneck of design iteration time.
To shrink the gap in performance modeling of integrated circuits (ICs), ML techniques are widely applied for fast circuit evaluation.

We discuss the studies based on whether their input features are extracted from pre-layout or post-layout information.
\textcircled{\small{1}}
Given design schematics, parasitics in layouts can be predicted from pre-layout stage, which helps bridge the gap of performance difference between pre-layout and post-layout simulations.
ParaGraph \cite{ren2020paragraph} builds a GNN model to predict layout-dependent parasitics and physical device parameters.
MLParest \cite{shook2020mlparest} shows that non-graph based methods (e.g., random forest) also work well for estimating interconnect parasitics, whereas the lack of placement information may cause large variations in predictions.
\textcircled{\small{2}}
Given circuit schematics as well as device information as inputs, it is possible to directly model post-layout performance from pre-layout designs.
Alawieh \textit{et al.} \cite{alawieh2017efficient} propose a hierarchical method that combines the Bayesian co-learning framework and semi-supervised learning to predict power consumption.
The entire circuit schematic is partitioned into multiple blocks to build block-level performance models, upon which circuit-level performance models are built. By combining these two low-dimensional models with a large amount of unlabeled data, pseudo samples can be labeled with almost no cost. Finally, a high-dimensional performance model mapping low-level features to circuit-level metrics is trained with pseudo samples and a small amount of labeled samples, which demonstrates the feasibility of performance modeling with inadequate labeled samples.
Several variants of Bayesian-based methods also perform well for estimating post-layout performance, e.g., combining Bayesian regression with SVM to predict circuit performance \cite{pan2019late} and using Bayesian DNNs to compare circuit designs \cite{hakhamaneshi2019bagnet}.
\textcircled{\small{3}}
Since post-layout simulations with SPICE-like simulators is time-consuming, ML techniques are applied to quickly assess layout design performance \cite{li2020exploring}.
To make better use of structural information inside layouts, intermediate layout placement results are represented as 3D images to feed a 3D CNN model \cite{liu2020towards}, or encoded as graphs to train a customized GNN model \cite{li2020customized}, with the goal to predict whether a design specification is satisfied.

\begin{table}[tbp]
\vspace{-10pt}
\caption{Summary of applying ML techniques for performance modeling and prediction in design automation.}
\vspace{-10pt}
\label{table:model_eda}
\centering
    \tiny
    \renewcommand{\arraystretch}{1}
    \setlength{\tabcolsep}{0.5pt}
\begin{tabular}{c|c|c|c}
\toprule
\textbf{Domain} & \textbf{Prediction of} & \textbf{Technique} & \textbf{Input} \\ \midrule
\multirow{5}{*}{\textbf{\begin{tabular}[c]{@{}c@{}}Analog \\ Circuit \\ $( \S$ \ref{sec:model_analog}$)$ \end{tabular}}} & Parasitics & GNN \cite{ren2020paragraph},  random forest \cite{shook2020mlparest} & Circuit schematics \\ \cline{2-4} 
 & Power/area/bandwidth & \begin{tabular}[c]{@{}c@{}}Bayesian co-learning and semi-supervised learning \cite{alawieh2017efficient}, \\ Bayesian regression and SVM \cite{pan2019late}\end{tabular} & \multirow{2}{*}{\begin{tabular}[c]{@{}c@{}}Circuit schematics \\ and device information\end{tabular}} \\ \cline{2-3}
 & \begin{tabular}[c]{@{}c@{}}Probability of superiority  between designs\end{tabular} & Bayesian DNN \cite{hakhamaneshi2019bagnet} &  \\ \cline{2-4} 
 & \begin{tabular}[c]{@{}c@{}}Gain/unity gain frequency/bandwidth/phase margin\end{tabular} & \begin{tabular}[c]{@{}c@{}}SVM/ANN/random forest \cite{li2020exploring}, 3D CNN \cite{liu2020towards}, GNN \cite{li2020customized}\end{tabular} & \multirow{2}{*}{Circuit placement} \\ \cline{2-3}
 & Electromagnetic properties & GNN \cite{zhang2019circuit} &  \\ \hline
\multirow{9}{*}{\makecell[c]{\textbf{HLS} \\ $( \S$ \textbf{\ref{sec:model_hls}}$)$}} & \begin{tabular}[c]{@{}c@{}}Area/latency/throughput/logic utilization\end{tabular} & \begin{tabular}[c]{@{}c@{}}Random forest  \cite{liu2013learning,meng2016adaptive}, transfer learning \cite{kwon2020transfer}\end{tabular} & Directives in HLS scripts \\ \cline{2-4} 
 & Resource utilization & ANN \cite{koeplinger2016automatic} & \multirow{3}{*}{IR graphs from HLS front-ends} \\ \cline{2-3}
 & Resource mapping and clustering & GraphSAGE \cite{ustun2020accurate} &  \\ \cline{2-3}
 & Routing  congestion & \begin{tabular}[c]{@{}c@{}}Linear regression/ANN/gradient boosted regression tree \cite{zhao2019machine}\end{tabular} &  \\ \cline{2-4} 
 & Power & \begin{tabular}[c]{@{}c@{}}Linear regression/SVM/tree-based models/DNNs \cite{lin2020hl}\end{tabular} & IR graphs and HLS reports \\ \cline{2-4} 
 & \begin{tabular}[c]{@{}c@{}}Throughput and throughput-to-area ratio\end{tabular} & \begin{tabular}[c]{@{}c@{}}Ensemble learning by stacked regression \cite{makrani2019pyramid}\end{tabular} & \multirow{2}{*}{HLS reports} \\ \cline{2-3}
 & Resource utilization and timing & \begin{tabular}[c]{@{}c@{}}Linear regression/ANN/gradient tree boosting \cite{dai2018fast}\end{tabular} &  \\ \cline{2-4} 
 & Cross-platform latency and power & Random forest \cite{o2018hlspredict} & CPU program counters \\ \cline{2-4} 
 & Speedup over an ARM processor & ANN \cite{makrani2019xppe} & \begin{tabular}[c]{@{}c@{}}Application characteristics, \\ HLS reports, FPGA configurations\end{tabular} \\ \hline
\multirow{8}{*}{\textbf{\begin{tabular}[c]{@{}c@{}}Logic and \\ Physical \\ Synthesis \\ $( \S$ \ref{sec:model_logic}$)$\end{tabular}}} & Area/delay & CNN \cite{yu2018developing}, LSTM \cite{yu2020decision} & Synthesis flows \\ \cline{2-4} 
 & \multirow{2}{*}{DRVs} & \begin{tabular}[c]{@{}c@{}}Linear regression/ANN/decision tree \cite{li2016machine}, MARS \cite{qi2014accurate}\end{tabular} & \begin{tabular}[c]{@{}c@{}}Placement and GR information\end{tabular} \\ \cline{3-4} 
 &  & MARS/SVM  \cite{chan2016beol}, MLP \cite{tabrizi2018machine} & Placement \\ \cline{2-4} 
 & \multirow{2}{*}{DRC hotspots} & FCN \cite{xie2018routenet} & \begin{tabular}[c]{@{}c@{}}Placement and GR information\end{tabular} \\ \cline{3-4} 
 &  & Variant of FCN \cite{liang2020drc} & \multirow{3}{*}{Placement} \\ \cline{2-3} 
 & GR congestion map & FCN \cite{chen2020pros} & \\ \cline{2-3} 
 & \multirow{2}{*}{Routing congestion in FPGAs} & Linear regression   \cite{maarouf2018machine} &  \\ \cline{3-4} 
 &  & Conditional GAN \cite{yu2019painting,alawieh2020high} & Post-placement images \\ \bottomrule
\end{tabular}
\vspace{-15pt}
\end{table}

\subsubsection{High-Level Synthesis (HLS)}
\label{sec:model_hls}
HLS is an automated transformation from behavioral languages (e.g., C/C++/SystemC) to register-transfer level (RTL) designs, which significantly expedites the development of hardware designs involving with field-programmable gate arrays (FPGAs) or application-specific integrated circuits (ASICs).
Since HLS tools usually take considerable time to synthesize each design, it prevents designers from exploring design space sufficiently, which motivates the application of ML models for fast and accurate performance estimation.

In performance estimation of HLS designs, the input features to ML models are extracted from three major sources: HLS directives, IRs from HLS front-ends, and HLS reports.
\textcircled{\small{1}} Taking the directives in an HLS script as input features, random forest is capable to forecast different design metrics, such as area and effective latency \cite{liu2013learning}, and throughput and logic utilization \cite{meng2016adaptive}.
In order to reuse knowledge from previous experiences, a transfer learning approach \cite{kwon2020transfer} can transfer knowledge across different applications or synthesis options.
\textcircled{\small{2}} Taking advantages of IR graphs generated by HLS front-ends, Koeplinger \textit{et al.} \cite{koeplinger2016automatic} count resource requirements of each node in graphs by using pre-characterized area models, which are then used as inputs to ANNs to predict the LUT routing usage, register duplication, and unavailable LUTs.
The exploitation of GNNs makes it possible to automatically predict the mapping from arithmetic operations in IR graphs to different resources on FPGAs \cite{ustun2020accurate}. 
To forecast post-implementation routing congestion, Zhao \textit{et al.} \cite{zhao2019machine} build a dataset that connects the routing congestion metrics after RTL implementation with operations in IRs, with the goal to train ML models locating highly congested regions in source code.
\textcircled{\small{3}} Taking the information that can be directly extracted from HLS reports, Dai \textit{et al.} \cite{dai2018fast} try several ML models (linear regression, ANN, and gradient tree boosting) to predict post-implementation resource utilization and timing. 
Pyramid \cite{makrani2019pyramid} applies the ensemble learning by stacked regression to accurately estimate the throughput or the throughput-to-area ratio.
HL-Pow \cite{lin2020hl} employs features from both IR graphs and HLS reports to predict the power by a variety of ML models.

The surge of heterogeneous platforms with FPGA/AISC and CPU provides more possibility of hardware/software co-design, motivating cross-platform performance predictions.
HLSPredict \cite{o2018hlspredict} uses random forest to predict FPGA cycle counts and power consumption based on program counter measurements obtained from CPU execution. 
While HLSPredict targets the same FPGA platform in training and testing, XPPE \cite{makrani2019xppe} considers different FPGA platforms, and uses ANNs to predict the speedup of an application on a target FPGA over an ARM processor.

\subsubsection{Logic and Physical Synthesis}
\label{sec:model_logic}
In digital design, logic synthesis converts RTL designs into optimized gate-level representations;
physical synthesis then transforms these design netlists into physical layouts.
Since these two stages may take hours or days to generate final bitstreams/layouts, many problems benefit from the power of ML models for fast performance estimation.

In logic synthesis, CNN models \cite{yu2018developing} or LSTM-based models \cite{yu2020decision} can be leveraged to forecast the delay and area after applying different synthesis flows on specific designs, where the inputs are synthesis flows represented in either matrices or time series.

In physical synthesis, routing is a sophisticated problem subject to stringent constraints, and EDA tools typically utilize a two-step method: global routing (GR) and detailed routing (DR). GR tool allocates routing resource coarsely, and provides routing plans to guide DR tools to complete the entire routing.
In general, routing congestion can be figured out during or after GR; the routability of a design is confirmed after DR and design rule checking (DRC).
Endeavors have been made to predict routability from early layout stages, so as to avoid excessive iterations back and forth between placement and routing.

In ASICs, some investigations predict routability by estimating the number of design rule violations (DRVs).
Taking GR results as inputs, Li \textit{et al.} \cite{li2016machine} explore several ML models (linear regression, ANN, and decision tree) to predict the number of DRVs, final hold slack, power, and area. 
Qi \textit{et al.} \cite{qi2014accurate} rely on placement data and congestion maps from GR as input features, and use a nonparametric regression technique, multivariate adaptive regression splines (MARS) \cite{friedman1991multivariate}, to predict the utilization of routing resource and the number of DRVs.
By merely leveraging placement information, it is possible to predict routability by MARS and SVM \cite{chan2016beol}, or to detect DR short violations by an MLP \cite{tabrizi2018machine}.
When representing placement information as images, fully convolutional networks (FCNs) are capable to predict locations of DRC hotspots by considering GR information as inputs \cite{xie2018routenet}, or to forecast GR congestion maps by formulating the prediction task as a pixel-wise binary classification using placement data \cite{chen2020pros}.
J-Net \cite{liang2020drc} is a customized FCN model, and takes both high-resolution pin patterns and low-resolution layout information from the placement stage as features to output a 2D array that indicates if the tile corresponding to each entry is a DRC hotspot.

In FPGAs, routing congestion maps can be directly estimated by linear regression \cite{maarouf2018machine} using feature vectors coming from pin counts and wirelength per area of SLICEs.
By constructing the routing congestion prediction as an image translation problem, a conditional GAN \cite{yu2019painting,alawieh2020high} is able to take post-placement images as inputs to predict congestion heat maps.
\section{ML as Design Methodology}
\label{sec:methodology}
This section introduces studies that directly employ ML techniques as the design methodology for computer architecture/systems.
Computer architecture and systems have been becoming increasingly complicated, making it expensive and inefficient for human efforts to design or optimize them.
In response, visionaries have argued that computer architecture and systems should be imbued with the capability to design and configure themselves, adjust their behaviors according to workloads' needs or user-specified constraints, diagnose failures, repair themselves from the detected failures, etc.
With strong learning and generalization capabilities, ML-based techniques are naturally suitable to resolve these considerations, which can adjust their policies during system designs according to long-term planning and dynamic workload behaviors.
As many problems in architecture/system design can be formulated as combinatorial optimization or sequential decision-making problems, RL is broadly explored and exploited.
Table \ref{table:dse_sys} and Table \ref{table:dse_eda} recapitulate the studies that apply ML techniques as the design methodology for computer architecture/system and design automation respectively, in terms of target tasks and adopted ML techniques.

\begin{table}[tbp]
\vspace{-10pt}
\caption{Summary of applying ML techniques as the design methodology for computer architecture/ systems.}
\vspace{-10pt}
\label{table:dse_sys}
\centering
    \tiny
    \renewcommand{\arraystretch}{0.9}
    \setlength{\tabcolsep}{3pt}
\begin{tabular}{c|c|c}
\toprule
\textbf{Domain} & \textbf{Task} & \textbf{Technique} \\ \midrule
\multirow{4}{*}{\textbf{\begin{tabular}[c]{@{}c@{}}Memory \\ System\\ Design \\ $( \S$ \ref{sec:dse_mem}$)$ \end{tabular}}} & Cache replacement policy & \begin{tabular}[c]{@{}c@{}}Perceptron learning   \cite{teran2016perceptron, jimenez2017multiperspective}, Markov decision process \cite{beckmann2017maximizing}, LSTM and SVM \cite{shi2019applying}\end{tabular} \\ \cline{2-3} 
 & Cache prefetching policy & \begin{tabular}[c]{@{}c@{}}Perceptron learning   \cite{wang2017data,bhatia2019perceptron}, contextual bandit \cite{peled2015semantic}, LSTM \cite{zeng2017long,hashemi2018learning,braun2019understanding,shineural}\end{tabular} \\ \cline{2-3} 
 & Memory controller policy & Q-learning   \cite{martinez2009dynamic,ipek2008self,mukundan2012morse} \\ \cline{2-3} 
 & Garbage collection & Q-learning \cite{kang2017reinforcement,kang2018dynamic} \\ \hline
\textbf{\begin{tabular}[c]{@{}c@{}}Branch\\ Prediction \\ $( \S$ \ref{sec:dse_bp}$)$\end{tabular}} & Branch direction & \begin{tabular}[c]{@{}c@{}}MLP \cite{calder1997evidence}, piecewise linear regression \cite{jimenez2005piecewise}, \\ perceptron   \cite{jimenez2001dynamic,jimenez2003fast,st2008low,jimenez2011optimized,jimenez2016multiperspective,garza2019bit}, CNN \cite{tarsa2019improving}\end{tabular} \\ \hline
\multirow{8}{*}{\makecell[c]{\textbf{NoC} \\ $( \S$ \textbf{\ref{NoC}}$)$}} & Link management & ANN   \cite{savva2012intelligent,reza2018neuro} \\ \cline{2-3} 
 & DVFS for routers & Q-learning   \cite{zheng2019energy,fettes2019dynamic} \\ \cline{2-3} 
 & Routing & Q-learning   \cite{boyan1994packet,kumar1997dual,majer2005packet,feng2010reconfigurable,ebrahimi2012haraq} \\ \cline{2-3} 
 & Arbitration policy & DQN   \cite{yin2018toward,yin2020experiences} \\ \cline{2-3} 
 & Adjusting injection rates & Q-learning \cite{daya2016quest},   ANN \cite{wang2019ann} \\ \cline{2-3} 
 & Selection of fault-tolerant modes & Q-learning   \cite{wang2019high,wang2019intellinoc} \\ \cline{2-3} 
 & Link placement in 3D NoCs & STAGE algorithm   \cite{das2015optimizing,das2016energy,joardar2018learning} \\ \cline{2-3} 
 & Loop placement in routerless NoCs & Advantage actor-critic with MCTS   \cite{lin2019optimizing} \\ \hline
\multirow{6}{*}{\textbf{\begin{tabular}[c]{@{}c@{}}Power \\ Management \\ $( \S$ \ref{power}$)$ \end{tabular}}} & DVFS and thread packing & Multinomial logistic regression   \cite{cochran2011pack} \\ \cline{2-3} 
 & DVFS and power gating & MLP \cite{ravi2017charstar} \\ \cline{2-3} 
 & \begin{tabular}[c]{@{}c@{}}DVFS, socket allocation, and use of HyperThreads\end{tabular} & Extra trees/gradient boosting/KNN/MLP/SVM \cite{imes2018energy} \\ \cline{2-3} 
 & \begin{tabular}[c]{@{}c@{}}DVFS for CPU cores/uncore/through-silicon interposers\end{tabular} & Propositional rule   \cite{aboughazaleh2007integrated}, ANN \cite{won2014up}, Q-learning   \cite{pd2015q} \\ \cline{2-3} 
 & \begin{tabular}[c]{@{}c@{}}DVFS for CPU-GPU heterogeneous platforms\end{tabular} & Weighted majority algorithm   \cite{ma2012greengpu} \\ \cline{2-3} 
 & DVFS for multi-/many-core systems & \begin{tabular}[c]{@{}c@{}}Q-learning \cite{bai2017voltage}, semi-supervised RL \cite{juan2012power}, hierarchical Q-learning   \cite{pan2014scalable,chen2015distributed,chen2017profit}\end{tabular} \\ \hline
\multirow{6}{*}{\textbf{\begin{tabular}[c]{@{}c@{}}Resource \\ Management \\ \& Task \\ Allocation \\ $( \S$ \ref{resource management}$)$\end{tabular}}} & Tuning architecture configurations & \begin{tabular}[c]{@{}c@{}}Maximum likelihood   \cite{dubach2010predictive}, statistical machine learning   \cite{ganapathi2009case,blanton2015statistical}\end{tabular} \\ \cline{2-3} 
 & Dynamic cache partitioning & Enforced subpopulations   \cite{gomez2001neuro}, Q-learning \cite{jain2016machine} \\ \cline{2-3} 
 & Task allocation in many-core systems & Q-learning   \cite{lu2015reinforcement}, DDPG \cite{wu2020core} \\ \cline{2-3} 
 & Workflow management & SVM and random forest   \cite{esteves2018adaptive} \\ \cline{2-3} 
 & Hardware resource assignment & REINFORCE   \cite{kao2020confuciux}, Bayesian optimization \cite{apollo} \\ \cline{2-3} 
 & Device placement & \begin{tabular}[c]{@{}c@{}}REINFORCE   \cite{mirhoseini2017device,mirhoseini2018hierarchical}, policy gradient \cite{addanki2018placeto}, PPO \cite{gao2018spotlight,zhou2019gdp}\end{tabular} \\ \hline
\multirow{2}{*}{\makecell[c]{\textbf{Scheduling}\\ $( \S$ \textbf{\ref{scheduling}}$)$}} & Scheduling jobs in single-core processors & Q-learning   \cite{whiteson2004adaptive} \\ \cline{2-3} 
 & Scheduling jobs in multi-processor systems & Value-based RL   \cite{fedorova2007operating,vengerov2009reinforcement} \\ \hline
\multirow{9}{*}{\textbf{\begin{tabular}[c]{@{}c@{}}Data Center\\ Management \\ $( \S$ \ref{data center}$)$\end{tabular}}} & Assignment of servers to applications & Value-based RL   \cite{tesauro2005online,tesauro2007reinforcement} \\ \cline{2-3} 
 & Content allocation in CDNs & Fuzzy RL   \cite{vengerov2002adaptive} \\ \cline{2-3} 
 & \begin{tabular}[c]{@{}c@{}}Placement of virtual machines onto physical machines\end{tabular} & PPO \cite{balajifireplace} \\ \cline{2-3} 
 & Traffic optimization & Policy gradient and DDPG   \cite{chen2018auto} \\ \cline{2-3} 
 & Scheduling jobs with complex dependency & REINFORCE \cite{mao2019learning} \\ \cline{2-3} 
 & Straggler diagnosis & Statistical ML   \cite{zheng2018hound} \\ \cline{2-3} 
 & Data-center-level caching policy & \begin{tabular}[c]{@{}c@{}}Decision tree \cite{wang2018efficient}, LSTM \cite{narayanan2018deepcache}, gradient boosting \cite{song2020learning}, DDPG \cite{wu2020phoebe}\end{tabular} \\ \cline{2-3} 
 & Bitrate selection for video chunks & A3C \cite{mao2017neural, yeo2018neural} \\ \cline{2-3} 
 & \begin{tabular}[c]{@{}c@{}}Scheduling video workloads in hybrid CPU-GPU clusters\end{tabular} & DQN \cite{zhang2018learning} \\ \hline
 \multirow{3}{*}{\textbf{\begin{tabular}[c]{@{}c@{}}Code\\ Generation \\ $( \S$ \ref{code_generation}$)$\end{tabular}}} & Code completion & N-gram model and RNN   \cite{raychev2014code} \\ \cline{2-3} 
 & Code   generation & LSTM  \cite{cummins2017synthesizing} \\ \cline{2-3} 
 & Program translation & Tree-to-tree encoder-decoder \cite{chen2018tree,fu2019coda}, seq2seq \cite{kim2019case}, transformer \cite{lachaux2020unsupervised} \\ \hline
\multirow{6}{*}{\makecell[c]{\textbf{Compiler}\\ $( \S$ \textbf{\ref{sec:compiler}}$)$}} & Instruction scheduling & Temporal difference   \cite{mcgovern2002building},  projective reparameterization \cite{jain2019learning} \\ \cline{2-3} 
 & Improving compiler heuristics & NEAT \cite{coons2008feature}, LSTM \cite{cummins2017end} \\ \cline{2-3} 
 & Ordering of optimizations & NEAT   \cite{kulkarni2012mitigating} \\ \cline{2-3} 
 & Automatic vectorization & Imitation learning \cite{mendis2019compiler} \\ \cline{2-3} 
 & \begin{tabular}[c]{@{}c@{}}Program transformation for approximate computing\end{tabular} & MLP \cite{esmaeilzadeh2012neural,yazdanbakhsh2015neural} \\ \cline{2-3} 
 & Compilation   for DNN workloads & PPO \cite{ahn2019reinforcement},   policy gradient \cite{khadka2020optimizing} \\ \bottomrule
\end{tabular}
\end{table}

\subsection{Memory System Design}
\label{sec:dse_mem}
The "memory wall" has been a performance bottleneck in von Neumann architectures, where computation is orders of magnitude faster than memory access.
To alleviate this problem, hierarchical memory systems are widely used and there arise optimizations for different levels of memory systems.
As both the variety and the size of modern workloads are drastically growing, conventional designs that are based on heuristics or intuitions may not catch up with the demand of the ever-growing workloads, leading to sharply degradation in performance.
As such, many studies resort to ML-based techniques to design smart and intelligent memory systems.

\subsubsection{Cache}
\label{cache}
The conspicuous disparity in latency and bandwidth between CPUs and memory systems motivates investigations for efficient cache management.
There are two major types of studies on cache optimization: improving cache replacement policies, and designing intelligent prefetching policies.
\textcircled{\small{1}} 
To develop cache replacement policies, perceptron learning is employed to predict whether to bypass or reuse a referenced block in the last-level cache (LLC) \cite{teran2016perceptron,jimenez2017multiperspective}.
Instead of using perceptrons, Beckmann \textit{et al.} \cite{beckmann2017maximizing} model the cache replacement problem as a Markov decision process and replace lines according to the difference between their expected hits and the average hits.
Shi \textit{et al.}~\cite{shi2019applying} train an attention-based LSTM model offline to extract insights from history program counters, which are then used to build an online SVM-based hardware predictor to serve as the cache replacement policy.
\textcircled{\small{2}} 
To devise intelligent prefetchers, Wang \textit{et al.} \cite{wang2017data} propose a prefetching mechanism that uses conventionally table-based prefetchers to provide prefetching suggestions and a perceptron trained by spatio-temporal locality to reject unnecessary prefetching decisions, ameliorating the cache pollution problem. 
Similarly, Bhatia \textit{et al.} \cite{bhatia2019perceptron} integrate a perceptron-based prefetching filter with conventional prefetchers, increasing the coverage of prefetches without hurting accuracy.
Instead of the commonly used spatio-temporal locality, a context-based memory prefetcher \cite{peled2015semantic} leverages the semantic locality that characterizes access correlations inherent to program semantics and data structures, which is approximated by a contextual bandit model in RL.
Interpreting semantics in memory access patterns is analogous to sequence analysis in natural language processing (NLP), and thus several studies use LSTM-based models and treat the prefetching as either a regression problem \cite{zeng2017long} or a classification problem \cite{hashemi2018learning}.
Even with better performance, especially for long access sequences and noise traces, LSTM-based prefetchers suffer from long warm-up and prediction latency, and considerable storage overheads.
The discussion of how hyperparameters impact LSTM-based prefetchers' performance \cite{braun2019understanding} highlights that the lookback size (i.e. memory access history window) and the LSTM model size strongly affect prefetchers' learning ability under different noise levels or workload patterns.
To accommodate the large memory space, Shi \textit{et al.} \cite{shineural} introduce a hierarchical sequence model to decouple predictions of pages and offsets by using two separate attention-based LSTM layers, whereas the corresponding hardware implementation is impractical for actual processors.

\subsubsection{Memory Controller}
Smart memory controllers can significantly improve memory bandwidth utilization.
Aiming at a self-optimizing memory controller adaptive to dynamically changing workloads, it can be modeled as an RL agent that always selects legal DRAM commands with the highest expected long-term performance benefits (i.e., Q-values) \cite{martinez2009dynamic,ipek2008self}.
To allow optimizations toward various objectives, this memory controller is then improved in two major aspects \cite{mukundan2012morse}. First, the rewards of different actions (i.e., legal DRAM commands) are automatically calibrated by genetic algorithms to serve different objective functions (e.g., energy, throughput, etc). Second, a multi-factor method that considers the first-order attribute interactions is employed to select proper attributes used for state representations.
Since both of them use table-based Q-learning and select limited attributes to represent states, the scalability may be a concern and their performance could be improved with more informative representations.


\subsubsection{Others}

A variety of work targets different parts of the memory system.
Margaritov \textit{et al.} \cite{margaritovvirtual} accelerate virtual address translation through learned index structures \cite{kraska2018case}. The results are encouraging in terms of the accuracy, which reaches almost 100\% for all tested virtual addresses; yet this method has unacceptably long inference latency, leaving practical hardware implementation as the future work.
Wang \textit{et al.} \cite{wang2016reducing} reduce data movement energy in interconnects by exploiting asymmetric transmission costs of different bits, where data blocks to be transmitted are dynamically grouped by K-majority clustering to derive energy-efficient expressions for transmission.
In terms of garbage collection in NAND flash, Kang \textit{et al.} \cite{kang2017reinforcement} propose an RL-based method to reduce the long-tail latency. The key idea is to exploit the inter-request interval (idle time) to dynamically decide the number of pages to be copied or whether to perform an erase operation, where decisions are made by table-based Q-learning. Their following work \cite{kang2018dynamic} considers more fine-grained states, and introduces a Q-table cache to manage key states among enormous amount of states.

\subsection{Branch Prediction}
\label{sec:dse_bp}
Branch predictor is one of the mainstays of modern processors, significantly improving the instruction-level parallelism. As pipelines gradually deepen, the penalty of mis-prediction increases. Traditional branch predictors often consider limited history length, which may hurt the prediction accuracy.
In contrast, the perceptron/MLP-based predictors can handle long histories with reasonable hardware budgets, outperforming prior state-of-the-art non-ML-based predictors.

Starting with a static branch predictor trained with static features from program corpus and control flow graphs, an MLP is used to predict the direction of a branch at compile time \cite{calder1997evidence}.
Later, a dynamic branch predictor uses a perceptron-based method \cite{jimenez2001dynamic}. It hashes the branch address to select the proper perceptron and computes the dot product accordingly to decide whether to take this branch, which shows great performance on linearly separable branches. 
Its latency and accuracy can be further improved by applying ahead pipelining and selecting perceptrons based on path history \cite{jimenez2003fast}.
To attain high accuracy in non-linearly separable branches, the perceptron-based prediction is generalized as piecewise linear branch prediction \cite{jimenez2005piecewise}.
In addition to the path history, multiple types of features from different organizations of branch histories can be leveraged to enhance the overall performance \cite{jimenez2016multiperspective}.
When considering practical hardware implementation of branch predictors, SNAP \cite{st2008low} leverages current-steering digital-to-analog converters to transfer digital weights into analog currents and replaces the costly digital dot-product computation to the current summation.
Its optimized version \cite{jimenez2011optimized} equips several new techniques, such as the use of global and per-branch history, trainable scaling coefficients, dynamic training thresholds, etc.

Rather than making binary decisions of whether to take a certain branch, it is possible to directly predict the target address of an indirect branch at the bit level via perceptron-based predictors \cite{garza2019bit}.
While high accuracy is achieved by current perceptron/MLP-based predictors, Tarsa \textit{et al.} \cite{tarsa2019improving} notice that a small amount of static branch instructions are systematically mispredicted, referred to as hard-to-predict branches (H2Ps). Consequently, they propose a CNN helper predictor for pattern matching of history branches, ultimately improving accuracy for H2Ps in conditional branches.

\subsection{NoC Design}
\label{NoC}
The aggressive transistor scaling has paved the way for integrating more cores in a single chip or processor.
With the increasing number of cores per chip, NoC plays a gradually crucial role, since it is responsible for inter-core communication and data movement between cores and memory hierarchies.
Several problems attracting attention are as follows. 
First, communication energy scales slower than computation energy \cite{borkar2013exascale}, implying necessity to improve power efficiency of NoCs. 
Second, the complexity of routing or traffic control grows with the number of cores per chip and this problem is even exacerbated by the rising variety and irregularity of workloads.
Third, with the continuous scaling down of transistors, NoCs are more vulnerable to different types of errors and thus reliability becomes a key concern.
Fourth, some non-conventional NoC architectures might bring promising potentials in the future, whereas they usually come with large design spaces and complex design constraints, which is nearly impossible for manually optimization.
Among all aforementioned fields, ML-based design techniques display their strength and charm.

\subsubsection{Link Management and DVFS}
Power consumption is one crucial concern in NoCs, in which links usually consume a considerable portion of network power.
While turning on/off links according to a static threshold of link utilization is a trivial way to reduce power consumption, it can not adapt to dynamically changing workloads.
Savva \textit{et al.} \cite{savva2012intelligent} use multiple ANNs for dynamic link management. Each ANN is responsible for one region of the NoC, and dynamically computes a threshold for every time interval to turn on/off links given the link utilization of each region. Despite significant power savings with low hardware overheads, this approach causes long latency in routing.
In order to meet certain power and thermal budgets, hierarchical ANNs \cite{reza2018neuro} are used to predict optimal NoC configurations (i.e., link bandwidth, node voltage and task assignment to nodes), where the global ANN predicts globally optimal NoC configurations exploiting local optimal energy consumption predicted by local ANNs.
To save dynamic power, several investigations \cite{zheng2019energy,fettes2019dynamic} employ per-router based Q-learning agents, which are offline trained ANNs to select optimal voltage/frequency levels for each router.

\subsubsection{Routing and Traffic Control}
With the increasing variety and irregularity of workloads and their traffic patterns, learning-based routing algorithms and traffic control approaches show superior performance due to their excellent adaptability.
\textcircled{\small{1}} 
As routing problems can be formulated as sequential decision-making processes, several studies apply Q-learning based approaches, namely the Q-routing algorithm \cite{boyan1994packet}, which uses local estimation of delivery time to minimize total packets delivery time, capable to handle irregular network topologies and keep a higher network load than the conventional shortest path routing.
Q-routing is then extended to several other scenarios, such as combining with dual RL to improve learning speed and routing performance \cite{kumar1997dual}, resolving packets routing in dynamic NoCs whose network structures/topologies are dynamically changing during runtime \cite{majer2005packet}, handling irregular faults in bufferless NoCs by the reconfigurable fault-tolerant Q-routing \cite{feng2010reconfigurable}, and enhancing the capability to reroute messages around congested regions by the congestion-aware non-minimal Q-routing \cite{ebrahimi2012haraq}.
In addition to routing problems, deep Q-network is also promising for NoC arbitration policies \cite{yin2018toward,yin2020experiences}, where the agent/arbiter grants a certain output port to the input buffer with the largest Q-value.
Even displaying some improvements in latency and throughput, the direct hardware implementation is impractical due to the complexity of deep Q-networks, and thus insights are distilled to derive a relatively simple circuitry implementation.
\textcircled{\small{2}} 
With the goal to control congestion in NoCs, the SCEPTER NoC architecture \cite{daya2016quest}, a bufferless NoC with single-cycle multi-hop traversals and a self-learning throttling mechanism, controls the injection of new flits into the network by Q-learning. Each node in the network independently selects whether to increase, decrease, or retain the throttle rate according to their Q-values, which conspicuously improves bandwidth allocation fairness and network throughput.
Wang \textit{et al.}~\cite{wang2019ann} design an ANN-based admission controller to determine the appropriate injection rate and the control policy of each node in a standard NoC.

\subsubsection{Reliability and Fault Tolerance}
With the aggressive technology scaling down, transistors and links in NoCs are more prone to different types of errors, indicating that reliability is a crucial concern and proactive fault-tolerant techniques are required to guarantee performance.
Wang~\textit{et al.}~\cite{wang2019high} employ per-router-based Q-learning agents to independently select one of four fault-tolerant modes, which can minimize the end-to-end packet latency and power consumption. These agents are pre-trained and then fine-tuned during runtime.
In their following work \cite{wang2019intellinoc}, these error-correction modes are extended and combined with various multi-function adaptive channel configurations, retransmission settings, and power management strategies, significantly improving latency, energy efficiency, and mean-time-to-failure.

\subsubsection{General Design}
With the growing number of cores per chip/system, the increasing heterogeneity of cores, and various performance targets, it is complicated to simultaneously optimize copious design knobs in NoCs.
One attempt to automated NoC design is the MLNoC \cite{rao2018mlnoc}, which utilizes supervised learning to quickly find near-optimal NoC designs under multiple optimization goals. MLNoC is trained by data from thousands of real-world and synthetic SoC (system-on-chip) designs, and evaluated with real-world SoC designs. Despite disclosure of limited details and absence of comprehensive comparison with other design methods, it shows superior performance to manually optimized NoC designs, delivering encouraging results.

Apart from conventional 2D mesh NoCs, a series of investigations focuses on 3D NoC designs, where the STAGE algorithm is applied to optimize vertical and planar placement of communication links in small-world network based 3D NoCs \cite{das2015optimizing,das2016energy}.
The STAGE algorithm repeatedly alternates between two stages, the base search that tries to find the local optima based on the learned evaluation function, and the meta-search that uses SVR to learn evaluation functions.
Later, the STAGE algorithm is extended for multi-objective optimization in heterogeneous 3D NoC systems \cite{joardar2018learning}, which jointly considers GPU throughput, average latency between CPUs and LLCs, temperature, and energy.
In terms of routerless NoCs that any two nodes are connected via at least one ring/loop, a deep RL framework that exploits Monte-Carlo tree search for efficient design space exploration is developed to optimize loop placements \cite{lin2019optimizing}, and the design constraints can be strictly enforced by carefully devising the reward function.

\subsection{Resource Allocation or Management}
\label{resource}
Resource allocation or management is the coordination between computer architecture/systems and workloads. Consequently, its optimization difficulty occurs with the booming complexity from both sides and their intricate interactions. 
ML-based approaches have blazed the trail to adjusting policies wisely and promptly pursuant to dynamic workloads or specified constraints.
\subsubsection{Power Management}
\label{power}
ML-based techniques have been applied broadly to improve power management, due to two main reasons.
First, power/energy consumption can be recognized as one metric of runtime costs.
Second, under certain circumstances there could be a hard or soft constraint/budget of power/energy, making power efficiency a necessity.

In consideration of power management for different parts of systems, PACSL \cite{aboughazaleh2007integrated} uses the propositional rule to adjust dynamic voltage scaling (DVS) for CPU cores and on-chip L2 cache, which achieves an improvement in the energy-delay product by 22\% on average (up to 46\%) over independently applying DVS for each part.
Won \textit{et al.} \cite{won2014up} coordinate an ANN controller with a proportional integral for uncore DVFS. The ANN controller can be either pre-trained offline by a prepared dataset or trained online by bootstrapped learning.
Manoj~\textit{et al.} \cite{pd2015q} deploy Q-learning to adaptively adjust the level of output-voltage swing at transmitters of 2.5D through-silicon interposer I/Os, under constraints of communication power and bit error rate.

From the system level, DVFS is one of the most prevalent techniques.
Pack \& Cap \cite{cochran2011pack} builds a multinomial logistic regression classifier that is trained offline and queried during runtime, to accurately identify the optimal operating point for both thread packing and DVFS under an arbitrary power cap.
GreenGPU \cite{ma2012greengpu} focuses on heterogeneous systems with CPUs and GPUs, and applies the weighted majority algorithm to scale frequency levels for both GPU cores and memory in a coordinated manner.
CHARSTAR \cite{ravi2017charstar} targets joint optimization of power gating and DVFS within a single core, where frequencies and configurations are dynamically selected by a lightweight offline trained MLP predictor.
To minimize energy consumption, Imes~\textit{et al.}~\cite{imes2018energy} use ML-based classifiers (e.g., extra trees, gradient boosting, KNN, MLP and SVM) to predict the most energy-efficient resource settings (specifically, tuning socket allocation, the use of HyperThreads, and processor DVFS) by using low-level hardware performance counters.
Bai~\textit{et al.}~\cite{bai2017voltage} consider the loss caused by on-chip regulator efficiency during DVFS, and try to minimize energy consumption under a parameterized performance constraint. The online control policy is implemented by a table-based Q-learning, which is portable across platforms without accurate modeling of a specific system.

A series of studies leverages RL for dynamic power management in multi-/many-core systems.
As systems scale up, these RL-based methods often suffer from state space explosion, and two types of methods are introduced to resolve the scalability issue.
\textcircled{\small{1}} 
By combining RL with supervised learning, a semi-supervised RL-based approach \cite{juan2012power} achieves linear complexity with the number of cores, which is able to maximize throughput ensuring power constraints and cooperatively control cores and uncores in synergy.
\textcircled{\small{2}} 
The exploitation of hierarchical Q-learning reduces the time complexity to $O(n \lg n)$, where $n$ denotes the number of cores.
Pan \textit{et al.} \cite{pan2014scalable} introduce multi-level Q-learning to select target power modes, where Q-values are approximated by a generalized radial basis function.
Table-based distributed Q-learning also performs well for DVFS \cite{chen2015distributed}, and there is one variant \cite{chen2017profit} aware of the priorities of different applications.

Some energy management policies target specific applications or platforms.
JouleGuard~\cite{hoffmann2015jouleguard} is a runtime control system coordinating approximate computing applications with system resource under energy budgets.
It uses a multi-arm bandit approach to identifying the most energy efficient system configuration, upon which application configurations are determined to maximize compute accuracy within energy budgets.
Targeting Intel SkyLake processors, a post-silicon CPU customization applies various ML models for dynamically clock-gating unused resource \cite{tarsa2019post}.

\subsubsection{Resource Management and Task Allocation}
\label{resource management}
Modern architectures and systems have been becoming so sophisticated and diverse that it is non-trivial to either optimize performance or fully utilize system resource.
This rapidly evolving landscape is further complicated by various workloads with specific requirements or targets.
In order to keep the pace, one cure is to develop more efficient and automated methods for resource management and task allocation, where ML-based techniques are excelled to explore large design spaces and simultaneously optimize multiple objectives, and preserve better scalablility and portability after carefully designed.

For a single-core processor, a regularized maximum likelihood approach \cite{dubach2010predictive} predicts the best hardware micro-architectural configuration for each phase of a program, based on runtime hardware counters. 
For multi-core processors, a statistical machine learning (SML) based method \cite{ganapathi2009case} can quickly find configurations that simultaneously optimize running time and energy efficiency. Since this method is agnostic to application and micro-architecture domain knowledge, it is a portable alternative to human expert optimization.
SML can also be applied as a holistic method to design self-evolving systems that optimize performance hierarchically across circuit, platform, and application levels \cite{blanton2015statistical}.
In addition to tuning architectural configurations, dynamic on-chip resource management is crucial for multi-core processors, where one example is dynamic cache partitioning.
In response to changing workload demands, an RNN evolved by the enforced subpopulations algorithm \cite{gomez2001neuro} is introduced to partition L2 cache dynamically.
When integrating dynamic partitioning of LLC with DVFS on cores and uncore, a co-optimization method using table-based Q-learning achieves much lower energy-delay products than any of the techniques applied individually \cite{jain2016machine}.

To guarantee efficient and reliable execution in many-core systems, task allocation should consider several aspects, such as heat and communication issues.
Targeting the heat interaction of processor cores and NoC routers, Lu \textit{et al.} \cite{lu2015reinforcement} apply Q-learning to assign tasks to cores based on current temperatures of cores and routers, such that the maximum temperature in the future is minimized.
Targeting the non-uniform and hierarchical on/off-chip communication capability in multi-chip many-core systems, core placement optimization \cite{wu2020core} leverages deep deterministic policy gradient (DDPG) \cite{lillicrap2015continuous} to map computation onto physical cores, able to work in a manner agnostic to domain-specific information.

Some studies pay attention to workflow management and general hardware resource assignment.
SmartFlux \cite{esteves2018adaptive} focuses on the workflow of data-intensive and continuous processing. It intelligently guides asynchronous triggering of processing steps with the help of predictions made by multiple ML models (e.g., SVM, random forest), which indicate whether to execute certain steps and to decide corresponding configurations upon each wave of data.
Given target DNN models, deployment scenarios, platform constraints, and optimization objectives (latency/energy), ConfuciuX~\cite{kao2020confuciux} applies a hybrid two-step scheme for optimal hardware resource assignments (i.e., assigning the number of processing elements and the buffer sizes to each DNN layer), where REINFORCE \cite{sutton2018reinforcement} performs a global coarse-grained search followed by a genetic algorithm for fine-grained tuning.
Apollo \cite{apollo} is a general architecture exploration framework for sample-efficient accelerator designs, which leverages ML-based black-box optimization techniques (e.g., Bayesian optimization) to optimize accelerator configurations to satisfy use-specified design constraints.

In heterogeneous systems with CPUs and GPUs, device placement refers to the process of mapping nodes in computational graphs of neural networks onto proper hardware devices.
Initially, computational operations are grouped manually, and assigned to devices by REINFORCE that employs a sequence-to-sequence RNN model as the parameterized policy \cite{mirhoseini2017device}.
Later, a hierarchical end-to-end model makes this manual grouping process automatic \cite{mirhoseini2018hierarchical}.
The training speed is further improved by introduction of proximal policy optimization (PPO) \cite{gao2018spotlight}.
Despite great advance brought by the above approaches, they are not transferable and a new policy should be trained from scratch specifically for each new computational graph.
By encode structure of computational graphs with static graph embeddings \cite{addanki2018placeto} or learnable graph embeddings \cite{zhou2019gdp}, the trained placement policy exhibits great generalizability to unseen neural networks.

\subsubsection{Scheduling}
\label{scheduling}
In classical real-time scheduling problems, the key task is to decide the order, according to which the currently unscheduled jobs should be executed by a single processor, such that the overall performance is optimized.
As multi-core processors have been the mainstream, the scheduling is gradually perplexing.
One major reason is that multiple objectives besides the performance should be carefully considered, such as balanced assignments among various cores and response time fairness.
Equipped with the capability to well understand the feedback provided by the environment and to dynamically adjust policies, RL is a common tool for real-time scheduling.

To optimize the execution order of jobs after they are routed to a single CPU core, Whiteson and Stone \cite{whiteson2004adaptive} propose an adaptive scheduling policy that exploits Q-routing, where the scheduler utilizes the router’s Q-table to assess a job’s priority and decides jobs' ordering accordingly so as to maximize the overall utility.
In multi-core systems, Fedorova~\textit{et al.} \cite{fedorova2007operating} present a blueprint for a self-tuning scheduling algorithm based on the value-based temporal-difference method in RL, aiming to maximize a cost function that is an arbitrary weighted sum of metrics of interest.
This algorithm is then improved to be a general method for online scheduling of parallel jobs \cite{vengerov2009reinforcement}, where the value functions are approximated by a parameterized fuzzy rulebase. This scheduling policy always selects to execute jobs with the maximum value functions in the job queue, which possibly preempts currently running jobs and squeezes some jobs into fewer CPUs than they ideally require, with the goal of achieving optimized long-term utility.

\subsection{Data Center Management}
\label{data center}
With the rapid scale expansion of data centers, issues that may be trivial in a single machine become increasingly challenging, let alone the inherently complicated problems.

Early work aims at a relatively simple scenario of resource allocation, i.e., to dynamically assign different numbers of servers to multiple applications. 
This problem can be modeled as an RL problem with service-level utility functions as rewards:
the arbiter will select a joint action that would bring the maximum total return after consulting local value functions estimated via either table-based methods \cite{tesauro2005online} or function approximation \cite{tesauro2007reinforcement}. 
In order to better model interactions among multiple agents, a multi-agent coordination algorithm with fuzzy RL \cite{vengerov2002adaptive} can be used to solve the dynamic content allocation in content delivery networks (CDNs), in which each requested content is modeled as an agent, trying to move toward the area with a high demand while coordinating with other agents/contents.
A recent innovation \cite{balajifireplace} pays attention to the placement of virtual machines onto physical machines, so as to minimize the peak-to-average ratio of resource usage across physical machines, where PPO and hindsight imitation learning are evaluated.

To improve data center performance and quality of experience (QoE) for users, ML-based techniques have been explored in a few directions.
\textcircled{\small{1}} It is important to efficiently schedule jobs and effectively diagnose stragglers within jobs.
Aiming at traffic optimization (e.g., flow scheduling, load balancing) in data centers, Chen~\textit{et al.} \cite{chen2018auto} develop a two-level RL system: peripheral systems, which are trained by DDPG, reside on end-hosts and locally make instant traffic optimization decisions for short flows; the central system, which is trained by policy gradient, aggregates global traffic information, guides behaviors of peripheral systems, and makes traffic optimization decisions for long flows.
Decima \cite{mao2019learning} exploits GNNs to represent cluster information and dependency among job stages, so that the RL-based scheduler can automatically learn workload-specific scheduling policies to schedule data processing jobs with complex dependency. 
Hound \cite{zheng2018hound} combines statistical ML with meta-learning to diagnose causes of stragglers at data-center-scale jobs.
\textcircled{\small{2}} It is essential to deploy an intelligent data-center-level cache.
DeepCache \cite{narayanan2018deepcache} employs an LSTM encoder-decoder model to predict future content popularity, which can be combined with existing cache policies to make smarter decisions.
Song \textit{et al.} \cite{song2020learning} apply gradient boosting machines to mimic a relaxed Belady algorithm that evicts an object whose next request is beyond a reuse distance threshold but not necessarily the farthest in the future.
Phoebe \cite{wu2020phoebe} is an online cache replacement framework leveraging DDPG to predict priorities of objects and to conduct eviction accordingly.
Considering non-history based features, Wang \textit{et al.}~\cite{wang2018efficient} build a decision tree to predict whether the requested file will be accessed only once in the future. These one-time-access files will be directly sent to users without getting into cache, to avoid cache pollution. 
\textcircled{\small{3}} From the workload perspective, video workloads on CDNs or clusters are prevalent but their optimization is quite challenging: first, network conditions fluctuate overtime and a variety of QoE goals should be balanced simultaneously; second, only coarse decisions are available and current decisions will have long-term effects on following decisions.
This scenario naturally matches the foundation of RL-based techniques.
To optimize users' QoE of streaming videos, adaptive bitrate algorithms have been recognized as the primary tool used by content providers, which are executed on client-side video players and dynamically choose a bitrate for each video chunk based on underlying network conditions.
Pensieve \cite{mao2017neural,yeo2018neural} applies asynchronous advantage actor-critic \cite{mnih2016asynchronous} to select proper bitrate for future video chunks based on resulting performance from past decisions.
When considering large-scale video workloads in hybrid CPU-GPU clusters, performance degradation often comes from uncertainty and variability of workloads, and unbalanced use of heterogeneous resources. 
To accommodate this, Zhang \textit{et al.} \cite{zhang2018learning} use two deep Q-networks to build a two-level task scheduler, where the cluster-level scheduler selects proper execution nodes for mutually independent video tasks and the node-level scheduler assigns interrelated video subtasks to appropriate computing units. This scheme enables the scheduling model to adjust policies according to runtime status of cluster environments, characteristics of video tasks, and dependency among video tasks.

\subsection{Code Generation and Compiler}
\label{sec:dse_code_compiler}
\subsubsection{Code Generation}
\label{code_generation}
Due to the similarities in syntax and semantics between programming languages and natural languages, the problem of code generation or translation is often modeled as an NLP problem or a neural machine translation (NMT) problem. 
Here, we would like to bring up a brief discussion. 
For more reference, a comprehensive survey \cite{allamanis2018survey} detailedly contrasts programming languages against natural languages, and discusses how these similarities and differences drive the design and application of different ML models in code.

Targeting code completion, several statistical language models (N-gram model, RNN, and a combination of these two) \cite{raychev2014code} are explored to select sentences that have the highest probability and satisfy constraints to fill up partial programs with holes.
As for code generation, CLgen \cite{cummins2017synthesizing} trains LSTM models by a corpus of hand-written code to learn semantics and structures of OpenCL programs, and generates human-like programs via iteratively sampling from the learned model.

Targeting program translation, NMT-based techniques are widely applied to migrate code from one language to another.
For example, a tree-to-tree model with the encoder-decoder structure effectively translates programs from Java to C\# \cite{chen2018tree};
the sequence-to-sequence (seq2seq) model can translate from CUDA to OpenCL \cite{kim2019case}.
Rather than translating between high-level programming languages, Coda \cite{fu2019coda} translates binary executables to the corresponding high-level code, which employs a tree-to-tree encoder-decoder structure for code sketch generation and an ensembled RNN-based error predictor for iterative error correction on the generated code.
Notably, these supervised NMT-based techniques may confront several issues: difficulty to generalize to programs longer than training ones, limited size of vocabulary sets, and scarcity of aligned input-output data.
Fully counting on unsupervised machine translation, TransCoder \cite{lachaux2020unsupervised} adopts a transformer architecture and uses monolingual source code to translate among C++, Java, and Python.

\subsubsection{Compiler}
\label{sec:compiler}
The complexity of compilers grows with the complexity of computer architectures and workloads. 
ML-based techniques can optimize compilers from many perspectives, such as instruction scheduling, compiler heuristics, the order to apply optimizations, hot path identification, auto-vectorization, and compilation for specific applications.
\textcircled{\small{1}} For instruction scheduling, the preference function of one scheduling over another can be computed by the temporal difference algorithm in RL \cite{mcgovern2002building}.
Regarding scheduling under highly-constrained code optimization, the projective reparameterization \cite{jain2019learning} enables automatic instruction scheduling under constraints of data-dependent partial orders over instructions.
\textcircled{\small{2}} For improving compiler heuristics, Neuro-Evolution of Augmenting Topologies (NEAT) \cite{coons2008feature} improves instruction placement heuristics by tuning placement cost functions.
To avoid manual feature engineering, LSTM-based model \cite{cummins2017end} automatically learns compiler heuristics from raw code, which constructs proper embeddings of programs and simultaneously learn the optimization process.
\textcircled{\small{3}} For choosing the appropriate order to apply different optimizations, NEAT \cite{kulkarni2012mitigating} can automatically generate beneficial optimization orderings for each method in a program.
\textcircled{\small{4}} For path profiling, CrystalBall \cite{zekany2016crystalball} uses an LSTM model to statically identify hot paths, the sequences of instructions that are frequently executed. As CrystalBall only relies on IRs, it avoids manual feature crafting and is independent of language or platform. 
\textcircled{\small{5}} For automatic vectorization, Mendis~\textit{et al.} \cite{mendis2019compiler} leverage imitation learning to mimic optimal solutions provided by superword-level-parallelism based vectorization \cite{mendis2018goslp}.
\textcircled{\small{6}} For compilation of specific applications, there are studies improving compilation for approximate computing or DNN applications.
Considering compilation for approximate computing, Esmaeilzadeh \textit{et al.} \cite{esmaeilzadeh2012neural} propose a program transformation method, which trains MLPs to mimic regions of approximable code and eventually replaces the original code with trained MLPs. The following work \cite{yazdanbakhsh2015neural} extends this algorithmic transformation to GPUs.
Considering compilation for DNNs, RELEASE \cite{ahn2019reinforcement} utilizes PPO to search optimal compilation configurations for DNNs.
EGRL \cite{khadka2020optimizing} optimizes memory placement of DNN tensors during compilation, which combines GNNs, RL, and evolutionary search to figure out optimal mapping onto different on-board memory components (i.e., SRAM, LLC, and DRAM).

\subsection{Chip Design and Design Automation}
\label{chip_design}
As technology scales down, the increased design complexity comes with growing process variations and reduced design margins, making chip design an overwhelmingly complex problem for human designers.
Recent advancements in ML create a chance to transform chip design workflows.

\subsubsection{Analog Design}
\label{sec:dse_analog}
Compared with the highly automated digital design counterpart, analog design usually demands many manual efforts and domain expertise.
First, analog circuits have large design spaces to search proper topology and device sizes. Second, there is an absence of a general framework to optimize or evaluate analog designs, and design specifications often vary case by case.
Recently, ML techniques have been introduced to expedite analog design automation.
We discuss these studies following the top-down flow of analog design: in the circuit level, a proper circuit topology is selected to satisfy system specifications; then in the device level, device sizes are optimized subject to various objectives. These two steps compose of pre-layout designs.
After circuit schematics are carefully designed, analog layouts in the physical level are generated.

\textcircled{\small{1}} 
In the circuit level, there is an attempt towards automatic circuit generation currently targeting two-port linear analog circuits \cite{rotman2020electric}.
The design specifications are encoded by a hypernetwork \cite{ha2016hypernetworks} to generate weights for an RNN model, which is trained to select circuit components and their configurations.
\textcircled{\small{2}} 
In the device level, the combination of RL and GNNs enables automatic transistor sizing \cite{wang2020gcn}, which is able to generalize across different circuit topologies or different technology nodes.
AutoCkt \cite{settaluri2020autockt} introduces transfer learning techniques into deep RL for automatic sizing, achieving 40$\times$ speedup over a traditional genetic algorithm.
Rosa \textit{et al.} \cite{rosa2019using} provide comprehensive discussions of how to address automatic sizing and layout of analog ICs via deep learning and ANNs.
\textcircled{\small{3}}
In the physical level, GeniusRoute \cite{zhu2019geniusroute} automates analog routing through the guidance learned by a generative neural network.
The analog placements and routing are represented as images to pass through a variational autoencoder (VAE) \cite{goodfellow2016deep} to learn routing likelihoods of each region. GeniusRoute achieves competitive performance to manual layouts and is capable to generalize to circuits of different functionality.
Liu \textit{et al.} \cite{liu2020closing} apply multi-objective Bayesian optimization to optimize combinations of net weighting parameters, which could significantly change floor plans and placement solutions, so as to improve analog layouts of building block circuits.

\begin{table}[tbp]
\vspace{-12pt}
\caption{Summary of applying ML techniques as the design methodology for design automation.}
\vspace{-10pt}
\label{table:dse_eda}
\centering
    \tiny
    \renewcommand{\arraystretch}{0.9}
    \setlength{\tabcolsep}{12pt}
\begin{tabular}{c|c|c|c}
\toprule
\multicolumn{2}{c|}{\textbf{Domain}} & \textbf{Task} & \textbf{Technique} \\ \midrule
\multirow{4}{*}{\textbf{\begin{tabular}[c]{@{}c@{}}Analog \\ Design \\ $( \S$ \ref{sec:dse_analog}$)$ \end{tabular}}} & \textbf{Circuit Level} & Generating circuit topology & RNN and hypernetwork   \cite{rotman2020electric} \\ \cline{2-4} 
 & \textbf{Device   Level} & Device sizing & Actor critic \cite{wang2020gcn,   settaluri2020autockt}, ANN \cite{rosa2019using} \\ \cline{2-4} 
 & \multirow{2}{*}{\textbf{Physical   Level}} & Routing & VAE \cite{zhu2019geniusroute} \\ \cline{3-4} 
 &  & Optimizing   layout configurations & Multi-objective Bayesian   optimization \cite{liu2020closing} \\ \cline{2-4} 
 \hline
\multirow{12}{*}{\textbf{\begin{tabular}[c]{@{}c@{}}Digital \\ Design \\ $( \S$ \ref{sec:dse_digital}$)$\end{tabular}}} & \multirow{3}{*}{\textbf{HLS}} & Optimizing loop unrolling pragma & Random forest   \cite{zacharopoulos2018machine} \\ \cline{3-4} 
 &  & Optimizing  placement of multiple pragmas & Bayesian optimization   \cite{mehrabi2020prospector} \\ \cline{3-4} 
 &  & Optimizing   resource pragma & Actor-critic and GNN   \cite{wu2021ironman} \\ \cline{2-4} 
 & \multirow{3}{*}{\textbf{\begin{tabular}[c]{@{}c@{}}Logic \\ Synthesis\end{tabular}}} & Selecting proper optimizers & MLP \cite{neto2019lsoracle} \\ \cline{3-4} 
 &  & Logic   optimization & Policy gradient   \cite{haaswijk2018deep}, actor-critic \cite{hosny2020drills} \\ \cline{3-4} 
 &  & Determining the maximum error of each node & Q-learning   \cite{pasandi2019approximate} \\ \cline{2-4} 
 & \multirow{5}{*}{\textbf{\begin{tabular}[c]{@{}c@{}}Physical \\ Synthesis\end{tabular}}} & Optimizing flip-flop placement in clock networks & K-means clustering   \cite{wu2016flip} \\ \cline{3-4} 
 &  & Optimizing   clock tree synthesis & Conditional GAN \cite{lu2019gan} \\ \cline{3-4} 
 &  & Optimizing   memory cell placement & PPO \cite{mirhoseini2020chip} \\ \cline{3-4} 
 &  & Optimizing   standard cell placement & Cast as an NN training problem   \cite{lin2019dreampiace} \\ \cline{3-4} 
 &  & Fix design rule violations & PPO \cite{rennvcell} \\ \bottomrule
\end{tabular}
\vspace{-12pt}
\end{table}

\subsubsection{Digital Design}
\label{sec:dse_digital}
For the studies applying ML techniques to directly optimize digital designs, we organize them following a top-down flow, i.e., HLS, logic synthesis, and physical synthesis.

The design space exploration in HLS designs usually relates to properly assigning directives (pragmas) in high-level source code, since directives significantly impact the quality of HLS designs by controlling parallelism, scheduling, and resource usage.
The optimization goal is often to find Pareto solutions between different objectives or to satisfy pre-defined constraints. 
With IR analysis, the employment of a random forest is able to select suitable loop unrolling factors to optimize a weighted sum of execution latency and resource usage \cite{zacharopoulos2018machine}.
Prospector \cite{mehrabi2020prospector} uses Bayesian optimization to optimize placement of directives (loop unrolling/pipelining, array partitioning, function inlining, and allocation), aiming to find Pareto solutions between execution latency and resource utilization in FPGAs. 
IronMan \cite{wu2021ironman} targets Pareto solutions between different resources while keeping the latency unchanged.
It combines GNNs with RL to conduct a finer-grained design exploration in the operation level, pursuing optimal resource allocation strategies by optimizing assignments of the resource pragma.

In logic synthesis, RTL-designs or logic networks are represented by directed acyclic graphs (DAGs). The goal is to optimize logic networks subject to certain constraints.
LSOracle \cite{neto2019lsoracle} employs an MLP to automatically decide which one of the two optimizers should be applied on different parts of circuits.
The logic optimization can be formulated as an RL problem solved by the policy gradient \cite{haaswijk2018deep} or the advantage actor-critic \cite{hosny2020drills}: the state is the current status of a design; the action is a transformation between two DAGs with equivalent I/O behaviors; the optimization objective is to minimize area or delay of designs.
Q-ALS \cite{pasandi2019approximate} aims at approximate logic synthesis and embeds a Q-learning agent to determine the maximum tolerable error of each node in a DAG, such that the total error rates at primary outputs are bounded by pre-specified constraints.

In physical synthesis, placement optimization is a popular topic.
\textcircled{\small{1}} 
To optimize flip-flop placement in clock networks, Wu \textit{et al.} \cite{wu2016flip} apply a modified K-means clustering to group post-placement flip-flops, and relocate these clusters by reducing the distance between flip-flops and their drivers while minimizing disruption of original placement results. 
To optimize clock tree synthesis (CTS), Lu \textit{et al.} \cite{lu2019gan} train a regression model that takes pre-CTS placement images and CTS configurations as inputs to predict post-CTS metrics (clock power, clock wirelength, and maximum skew), which is used as the supervisor to guide the training of a conditional GAN, such that the well-trained generator can recommend CTS configurations leading to optimized clock trees. 
\textcircled{\small{2}} 
Aiming at cell placement, a deep RL approach \cite{mirhoseini2020chip} is introduced to place macros (memory cells), after which standard cells are placed by a force-directed method. 
This method is able to generalize to unseen netlists, and outperforms RePlAce \cite{cheng2018replace} yet several times slower.
DREAMPlace \cite{lin2019dreampiace} casts the analytical standard cell placement optimization into a neural network training problem, achieving over 30$\times$ speedup without quality degradation compared to RePlAce.
NVCell \cite{rennvcell} is an automated layout generator for standard cells, which employs RL to fix DRVs after placement and routing.
\textcircled{\small{3}} 
ML-based techniques demonstrates their versatility in many design automation tasks, such as post-silicon variation extraction by sparse Bayesian learning, and post-silicon timing tuning to mitigate the effects caused by process variation \cite{zhuo2017accelerating}.
\section{Discussion and Potential Directions}
\label{sec:dis}
In this section, we discuss limitations and potentials of ML techniques for computer architecture and systems, which span the entire development and deployment stack that involves data, algorithms, implementation, and targets.
We also envision that the application of ML techniques could be the propulsive force for \textit{hardware agile development}.

\vspace{-5pt}
\subsection{Bridging Data Gaps}
Data are the backbone to ML, however, perfect datasets are sometimes non-available or prohibitively expensive to obtain in computer architecture and system domain.
Here, we would like to scrutinize two points, the gap between small data and big data, and non-perfect data.
\textcircled{\small{1}} 
In some EDA problems, such as placement and routing in physical synthesis, the simulation or evaluation is extremely expensive \cite{yanghua2016improving}, leading to data scarcity. As ML models usually require enough data to learn underlying statistics and make decisions, this gap between small data and big data often limits the capability of ML-based techniques. There have been different attempts to bridge this gap. From the algorithm side, algorithms that can work with small data await to be developed, where one current technique is Bayesian optimization that is effective in small parameter space \cite{khailany2020accelerating}; active learning \cite{settles2009active}, which significantly improve sample efficiency, may also be a cure to this problem. From the data side, generative methods can be used to generate synthetic data \cite{ding2019generative}, mitigating data scarcity.
\textcircled{\small{2}} 
Regarding non-perfect data, even if some EDA tools produce a lot of data, they are not always properly labeled nor presented in the form suitable to ML models. In the absence of perfectly labeled training data, possible alternatives are to use unsupervised learning, self-supervised learning \cite{hendrycks2019using}, or to combine supervised with unsupervised techniques \cite{alawieh2017efficient}. Meanwhile, RL could be a workaround where training data can be generated on-the-fly.

\vspace{-5pt}
\subsection{Developing Algorithms}
Despite the current achieved accomplishments, we are still expecting novel ML algorithms or schemes to further improve system modeling and optimization, with respect scalability,  domain knowledge interpretability, etc.

\textbf{New ML Schemes.}
Classical analytic-based methods usually adopt a bottom-up or top-down procedure, encouraging ML-based techniques to distill hierarchical structures of systems/architecture.
One example is hierarchical RL \cite{kulkarni2016hierarchical} that has flexible goal specifications and learns goal-directed behaviors in complex environments with sparse feedback. Such kind of models enables more flexible and effective multi-level design and control.
Additionally, many system optimizations involve participation of multiple agents, such as NoC routing, which are naturally suitable to the realm of multi-agent RL \cite{zhang2019multi}. These agents can be fully cooperative, fully competitive, or a mix of the two, enabling versatility of system optimization.
Another promising approach is self-supervised learning \cite{hendrycks2019using}, beneficial in both improving model robustness and mitigating data scarcity.
While applying a single ML method solely has led to powerful results, hybrid methods, i.e., combining different ML techniques or combining ML techniques with heuristics, unleash more opportunities. For example, RL can be combined with genetic algorithms for hardware resource assignment \cite{kao2020confuciux}.

\textbf{Scalability.}
The system scaling-up poses challenges on scalability issues.
From the algorithm side, multi-level techniques can help reduce the computation complexity, e.g., multi-level Q-learning for DVFS \cite{pan2014scalable,chen2015distributed,chen2017profit}.
One implicit workaround is to leverage transfer learning:
the pre-training is a one-time cost, which can be amortized in each future use; the fine-tuning provides flexibility between a quick solution from the pre-trained model and a longer yet better one for a particular task. 
Several examples \cite{wang2020gcn,settaluri2020autockt,mirhoseini2020chip} are discussed in Section \ref{chip_design}.

\textbf{Domain Knowledge and Interpretability.}
Making better use of domain knowledge unveils possibilities to choose more proper models for different system problems and provide more intuitions or explanations of why and how these models work.
By making analogy of semantics between memory access patterns/program languages and natural languages, the prefetching or code generation problems can be modeled as NLP problems, as discussed in Section \ref{cache} and Section \ref{code_generation}. 
By making analogy of graphical representations in many EDA problems, where data are intrinsically presented as graphs (e.g., circuits, logic netlists or IRs), GNNs are expected to be powerful in these fields \cite{khailany2020accelerating}. 
Several examples are provided in Section \ref{sec:model_eda} and Section \ref{chip_design}.

\subsection{Improving Implementations and Deployments}
To fully benefit from ML-based methods, we need to consider practical implementations, appropriate selection of deployment scenarios, and post-deployment model maintenance.

\textbf{Better Implementations.}
To enable practical implementations of ML-based techniques, improvement can be made from either the model side or software/hardware co-design \cite{sze2017efficient}.
From the model level, network pruning and model compression reduce the number of operations and model size \cite{han2016deep}; weight quantization improves computation efficiency by reducing the precision of operations/operands \cite{jacob2018quantization}.
From the co-design level, strategies that have been used for DNN acceleration could also be used in applying ML for system.

\textbf{Appropriate Scenarios: online vs. offline.}
When deploying ML-based techniques for system designs, it is crucial to deliberate design constraints under different scenarios.
Generally, existing work falls into two categories. 
\textcircled{\small{1}} 
ML-based techniques are deployed online or during runtime, no matter the training phase is performed online or offline. Obviously, the model complexity and runtime overhead are often strictly limited by specific constraints, e.g., power/energy, timing/latency, area, etc. 
To take one more step, if the online training/learning is further desired, the design constraint will be more stringent. One promising approach is to employ semi-online learning models, which have been applied to solve some classical combinatorial optimization problems, such as bipartite matching \cite{kumar2019semi} and caching \cite{kumar2020interleaved}. These models enable smooth interpolation between the best possible online and offline training algorithms.
\textcircled{\small{2}} ML-based techniques are applied offline, which often refers to architectural design space exploration. Such problems leverage ML-based techniques to guide system implementation, and once the designing phase is completed, ML models will not be invoked again. Thus, the offline applications can tolerate relatively higher overheads.

\textbf{Model Maintenance.}
In the case of offline training and online deployment, ML models employed for computer architecture domain, as in other scenarios, require regular maintenance and updating to meet performance expectations,
since workload variations over time and hardware aging often cause data drift or concept drift \cite{tsymbal2004problem}.
To proactively circumvent performance degradation of ML models, some measures could be taken during post-deployment periods.
\textcircled{\small{1}}
ML models can be retrained either at a regular interval or when key performance indicators are below certain thresholds. Retraining models regularly, regardless of their performance, is a more direct way, but it requires a clear understanding of how frequently a model should be updated under its own scenario. The model performance will decline if retraining intervals are too spaced out in the interim. Monitoring key performance indicators relies on a comprehensive panel of measurements that explicitly demonstrate model drift, whereas this may introduce additional hardware/software overhead and incorrect selection of measurements often defeats the intention of this method.
\textcircled{\small{2}} During the retraining of ML models, there is often a trade-off between newly collected data and previous data. Properly assigning importance of input data would improve retraining efficacy \cite{byrd2019effect}.

\vspace{-5pt}
\subsection{Supporting Non-homogeneous Tasks}
ML-based techniques are supposed to be applicable in both current architectures and emerging systems, leading to long-term advancement in computer architecture and systems.

\textbf{Non-homogeneous Components.}
Design and development for computer architectures are often based upon earlier-generation architectures of similar purpose, but commonly rely on next-generation hardware components that were not present in earlier generations. Examples include employment of new device nodes with technology scaling, and replacement of conventional constituents in memory systems with NVM- or PIM-based components.
In addition to the heterogeneity of components from different generations, one architecture or system usually consists of both standard parts from library and specialized/customized hardware components.
This provides the motivation that ML-assisted architectures/systems should have the flexibility to transfer among different-generation components, and to support standard and specialized parts simultaneously.

\textbf{Non-homogeneous Applications.}
In computer architecture and system design, some issues are universal, while others may arise with the advent of new architecture/systems and new workloads.
\textcircled{\small{1}} For evergreen design areas, several examples include caching in hardware/software/data centers (Section \ref{cache} and Section \ref{data center}), resource management and task allocation in single/multi-/many-core CPUs and heterogeneous systems (Section \ref{resource}), NoC design under various scenarios (Section \ref{NoC}), etc.
\textcircled{\small{2}} For problems aroused from new systems/workloads, transfer learning and meta-learning \cite{nichol2018first, vanschoren2018meta} could be helpful in either exploring new heuristics or directly deriving design methodology. 
For example, combining meta-learning with RL \cite{finn2017model} allows training a "meta" agent that is designed to adapt to a specific workload with only a few observations.

\vspace{-5pt}
\subsection{Facilitating General Tool Design and Hardware Agile Development}
Even though ML-based modeling significantly reduces the evaluation cost during design iteration, making great strides towards the landing of hardware agile development, there is still a long way to go in the ML-based design methodology prospective.
One ultimate goal might be the fully automated design, which should entangle two core capabilities: holistic optimization in system-wise, and easy migration across different systems, to enable rapid and agile hardware design.

\textbf{Holistic Optimization.}
Fueled by recent advancements, ML techniques have been increasingly explored and exploited in computer system design and optimization \cite{dean20201}. 
The target problems that await further endeavors could be multi-objective optimizations under highly constrained situations, or optimizing several components in a system simultaneously.
We envisage an ML-based system-wise and holistic framework with a panoramic vision: it should be able to leverage information from different levels of systems in synergy, so that it could thoroughly characterize system behaviors as well as their intrinsically hierarchical abstractions; it should also be able to make decisions in different granularity, so that it could control and improve systems precisely and comprehensively.

\textbf{Portable, Rapid, and Agile.}
Striving for portable, rapid, and agile hardware design, there are two potential directions.
\textcircled{\small{1}} The well-designed interfaces between systems/architectures and ML-based techniques would facilitate the portability across different platforms, since ML models can perform well without explicit descriptions of the target domain.
\textcircled{\small{2}} The proliferation of ML-based techniques have more or less transformed the workflow of design automation, directly driving rapid and agile hardware design.
We expect GNNs make better use of naturally graphical data in EDA field; we expect deep RL be a powerful and general-purpose tool for many EDA optimization problems, especially when the exact heuristic or objective is obscure; we expect these ML-based design automation tools enhance designers' productivity and thrive in the community.


\vspace{-5pt}
\section{Conclusion}
\label{sec:con}
The flourishing of ML would be retarded without the great systems and powerful architectures supportive to run these algorithms at scale. 
Now, it is the time to return the favor and let ML transform the way that computer architecture and systems are designed.
Existing work that applies ML for computer architecture/systems roughly falls into two categories: ML-based fast modeling that involves performance metrics or some other criteria of interest, and ML-based design methodology that directly leverages ML as the design tool.
We hope to see the virtuous cycle, in which ML-based techniques are efficiently running on the most powerful computers with the pursuit of designing the next generation computers. We hope ML-based techniques could be the impetus to the revolution of computer architecture and systems.

\vspace{-5pt}

\bibliographystyle{ACM-Reference-Format}
\bibliography{ref}

\end{document}